\documentclass{article}

\usepackage{arxiv}

\usepackage[utf8]{inputenc} 
\usepackage[T1]{fontenc}    
\usepackage{hyperref}       
\usepackage{url}            
\usepackage{booktabs}       
\usepackage{amsfonts}       
\usepackage{nicefrac}       
\usepackage{microtype}      
\usepackage{lipsum}		
\usepackage{graphicx}
\usepackage[round]{natbib}
\usepackage{doi}
\usepackage{amsmath}

\usepackage[dvipsnames]{xcolor}
\usepackage[normalem]{ulem}

\let\cite\citep

\title{Optimal Learning Rate Schedule for Balancing Effort and Performance}

\author{ \href{https://orcid.org/0009-0001-1742-004X}{\includegraphics[scale=0.06]{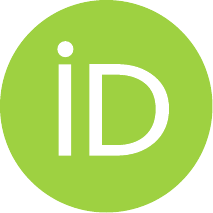}\hspace{1mm}Valentina Njaradi} \thanks{Equal Contribution.} \\
	Gatsby Computational Neuroscience Unit\\
	University College London\\
	\texttt{valentina.njaradi.23@ucl.ac.uk} \\
	\And
	\href{https://orcid.org/0000-0003-4673-8791}{\includegraphics[scale=0.06]{orcid.pdf}\hspace{1mm}Rodrigo Carrasco-Davis}$^*$  \\
	Gatsby Computational Neuroscience Unit\\
	University College London\\
    Princeton Neuroscience Institute\\
    \texttt{rodrigo.cd@princeton.edu} \\
    \And 
    \href{https://orcid.org/0000-0001-8713-9328}{\includegraphics[scale=0.06]{orcid.pdf}\hspace{1mm}Peter Latham}\\
	Gatsby Computational Neuroscience Unit\\
	University College London\\
    \And
    \href{https://orcid.org/0000-0002-9831-8812}
    {\includegraphics[scale=0.06]{orcid.pdf}\hspace{1mm}Andrew Saxe} \\
	Gatsby Computational Neuroscience Unit\\
	Sainsbury Wellcome Centre\\
	University College London \\
}
\date{}

\renewcommand{\headeright}{}
\renewcommand{\undertitle}{}
\renewcommand{\shorttitle}{Optimal Learning Rate Schedule for Balancing Effort and Performance}

\newcommand{\w}{\mathbf{w}}
\newcommand{\x}{\mathbf x}
\newcommand{\y}{\mathbf{y}}

\begin{document}
\maketitle

\begin{abstract}
   Learning how to learn efficiently is a fundamental challenge for biological agents and a growing concern for artificial ones. To learn effectively, an agent must regulate its learning speed, balancing the benefits of rapid improvement against the costs of effort, instability, or resource use. We introduce a normative framework that formalizes this problem as an optimal control process in which the agent maximizes cumulative performance while incurring a cost of learning. From this objective, we derive a closed-form solution for the optimal learning rate, which has the form of a closed-loop controller that depends only on the agent’s current and expected future performance. Under mild assumptions, this solution generalizes across tasks and architectures and reproduces numerically optimized schedules in simulations. In simple learning models, we can mathematically analyze how agent and task parameters shape learning-rate scheduling as an open-loop control solution. Because the optimal policy depends on expectations of future performance, the framework predicts how overconfidence or underconfidence influence engagement and persistence, linking the control of learning speed to theories of self-regulated learning. We further show how a simple episodic memory mechanism can approximate the required performance expectations by recalling similar past learning experiences, providing a biologically plausible route to near-optimal behaviour. Together, these results provide a normative and biologically plausible account of learning speed control, linking self-regulated learning, effort allocation, and episodic memory estimation within a unified and tractable mathematical framework.
\end{abstract}

Humans and other animals constantly face the problem of allocating their limited cognitive resources to learning. This means not only deciding what is worth learning, but also how to approach each learning task. For instance, when learning piano, one might take a quick learning strategy and memorize sequences of keys to play simple songs, while another might take a slower learning strategy, study music theory and learn to read sheet music to later generalize to a broader repertoire.  In this work, we focus on an abstract version of this general problem: how should learning systems regulate their learning speed when balancing performance against the costs of learning fast.

From childhood onward, humans actively regulate their learning \citep{taffoni_development_2014, raz_learning_2020, van_gog_role_2020}. They seek out information, select inputs that match their abilities \cite{kidd_goldilocks_2012}, and weigh the potential benefits of learning against its costs \cite{bonawitz_choosing_2018}. Adults, too, learn more effectively when they control the pace and structure of their study, compared to passive conditions \cite{castro_human_2008, markant_is_2014, coenen_asking_2019, tullis_effectiveness_2011}. One's own metacognitive control is fundamental: in studies of massed versus spaced learning, participants only benefited from spacing when it was aligned with their chosen learning plans \cite{son_metacognitive_2010}. Theories differ on which internal states drive such control: curiosity, boredom, and other motivational signals have all been implicated \cite{bazhydai_curiosity_2020, hidi_interest_2019, schwartenbeck_computational_2019, geana_boredom_2016, agrawal_temporal_2022}. But the underlying question remains: what computations guide decisions about how to learn?

Many factors shape learning schedules. Returning to the piano example, the “right” strategy depends on available practice time, effort, prior knowledge, task difficulty, the long-term value of playing an instrument, and even personal enjoyment. Evidence supports the importance of each, but studying them in isolation can yield conflicting results. For example, people sometimes favour intermediate and high-difficulty tasks \cite{baranes_effects_2014, schulz_structured_2019}, while other work identifies a preference for the “Goldilocks zone” where tasks are neither too easy nor too hard \cite{kidd_goldilocks_2012, wilson_eighty_2019, ten_humans_2021, cubit_visual_2021}. Time pressure also matters: when rushed, learners prioritize easy tasks, but given ample study time they choose harder ones \cite{son_metacognitive_2000}. Learners are also sensitive to gaps in their knowledge and prefer content they have not yet mastered \cite{kidd_goldilocks_2012, de_eccher_childrens_2024}. The value of learning in terms of future rewards \cite{masis_strategically_2023, obando_learning_2025} further influences choices, where rodents, monkeys and humans have been shown to weigh short-term gains against the long-term value of learning.

Along with all of its benefits, learning also comes with costs. Some are metabolic: human studies show links between metabolism and increased memory and learning functions \cite{potter_metabolic_2010, smith_glucose_2011, klug_learning_2022}, while in flies and butterflies, intensive learning reduces reproductive fitness and shortens lifespan \cite{mery_operating_2004, snell-rood_reproductive_2011, mery_cost_2005}. Others arise from the exertion of cognitive control itself. Scheduling and regulating learning is a form of cognitive control \cite{masis_value_2021, masis_strategically_2023, masis_learning_2024, carrasco-davis_meta-learning_2024}, which comes with its own costs that participants take into account when allocating control \cite{kool_mental_2018, kurzban_opportunity_2013, shenhav_expected_2013, shenhav_toward_2017, agrawal_temporal_2022}. When the cost of control is considered in a learning context, studies show that subjects often choose more cognitively demanding tasks over easier ones \cite{jarvis_effort_2022}, particularly when those tasks offer opportunities for learning \cite{sayali_learning_2023}. This suggests that humans can weigh the immediate costs of engaging in challenging tasks against the anticipated future benefits 
of learning \cite{obando_learning_2025}.

These findings suggest that humans continuously balance costs and benefits when deciding how to learn. Framed computationally, the problem of optimally regulating an agent's learning, so as to improve speed or quality, has been extensively studied in the field of meta-learning \cite{hospedales_meta-learning_2020}. In neural networks, meta-learning can take many forms: designing optimal learning curricula \cite{zhang_curriculum-based_2021, stergiadis_curriculum_2021, soviany_curriculum_2022}, finding optimal initial weights for a network \cite{finn_model-agnostic_2017}, continual learning \cite{parisi_continual_2019, wang_comprehensive_2024}, optimizing hyperparameters \cite{franceschi_bilevel_2018, baik_meta-learning_2020}, or determining optimal learning rate schedules throughout training \cite{nakamura_learning-rate_2021, mori_optimal_2025, mignacco_statistical_2025}. Still, designing an optimal learning speed schedule is computationally challenging, as in most cases it requires integrating and differentiating performance throughout the entire learning trajectory \cite{franceschi_bilevel_2018, carrasco-davis_meta-learning_2024, mori_optimal_2025, mignacco_statistical_2025}. For biological agents, this computation can be even more challenging, where the capacity for calculation of optimal control may be limited, and the necessary quantities may not be easily available. Nonetheless, managing learning is a key factor determining long-term success. All humans and animals face this problem on a daily basis, raising a key question: how do they do it?

In this paper, we take a normative approach to this problem by formalizing learning speed control as a trade-off between task performance and effort (\autoref{sec:the_framework}). We derive a closed-loop analytical expression for the optimal learning speed that depends only on quantities that could be available to the agent through learning, such as its current performance and its expected final performance, thereby avoiding the need to predict full learning trajectories (\autoref{sec:closed_loop}). We show that this solution generalizes across tasks and network architectures, compare it to suboptimal strategies, and explore how biological agents might learn and approximate their own expected final performance using memories of past learning experiences (\autoref{sec:estimating_final_performance}). For simple models like the perceptron, we derive exact open-loop solutions that provide intuition for how task difficulty and effort costs shape learning (\autoref{sec:solution_perceptron}). We extend the framework to settings with discounting of future rewards, presenting approximations for simple models and simulations that highlight the broader effects of discounting on optimal learning strategies (\autoref{sec:discount_factor_effects}). We compare results from our framework with animal experiments, and outline further experimental predictions (\autoref{sec:animal_behaviour}). Finally, we discuss broader implications of our theory for machine learning and cognitive neuroscience (\autoref{sec:discussion}).

\begin{figure}
    \centering
    \includegraphics[width=0.5\linewidth]{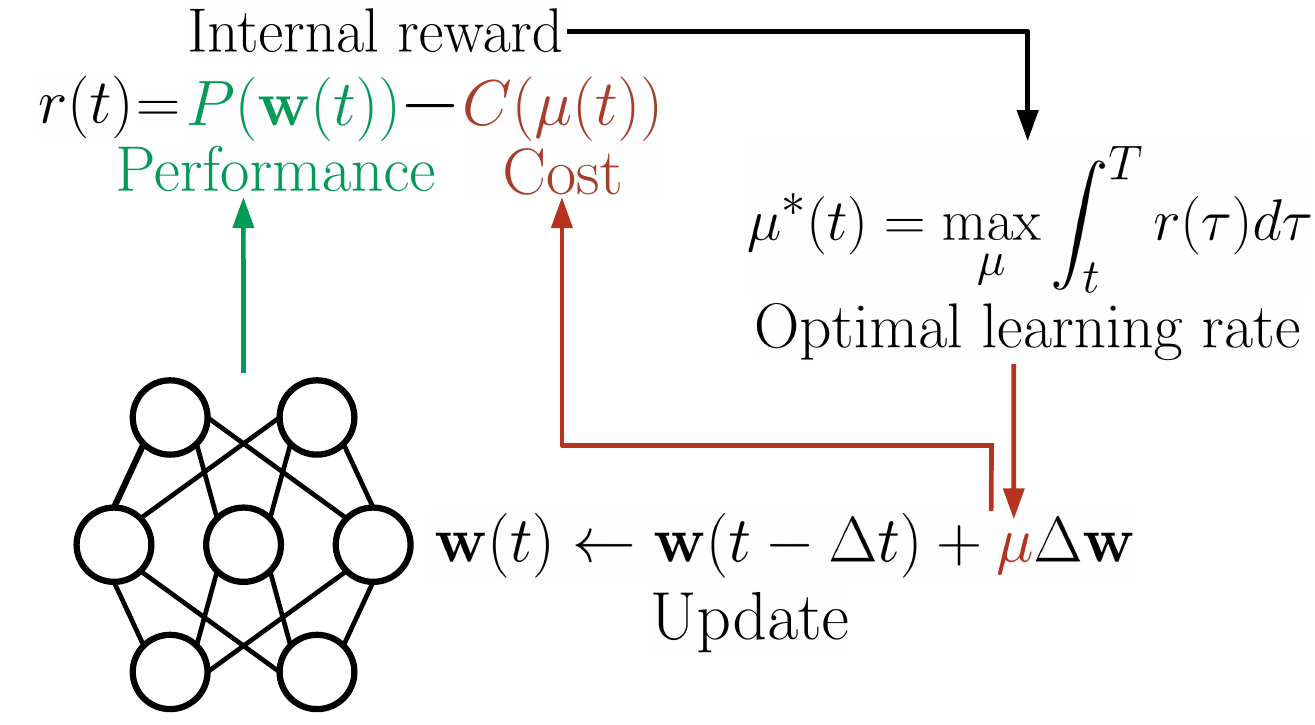}
    \caption{\textbf{Theoretical Framework}: During neural network training, the optimal learning rate for weight updates is computed to maximize the cumulative internal reward throughout learning. The internal reward is the difference between the reward collected through performing a task $P(t)$ and the cost $C(\mu(t))$ of learning with the rate $\mu$.}
    \label{fig:setup} 
\end{figure}

\section{Balancing Learning Effort and Performance} \label{sec:the_framework}

In our work, an agent learns a task during a learning episode of length $T$, and collects rewards throughout the episode while simultaneously incurring a cost of learning. The aim of the agent is to maximize its cumulative reward throughout learning while minimizing the effort required to do so. The agent can achieve this by adjusting its learning speed $\mu(t)$ throughout the episode (Fig.~\ref{fig:setup}). The agent's performance on the task throughout the learning episode is defined by a differentiable function $P(t)$, which depends on the parameters of the agent and the task. To apply the analytical methods described below, this function must be deterministic and continuously differentiable with respect to the agent’s parameters.

We model the agent as an artificial neural network with $N$ learnable parameters, or weights, collected in the vector $\textbf{w} \in \mathbb{R}^N$. In this case, $P(t)$ depends on time only through these weights: $P(t) = P(\mathbf{w}(t))$. The network's architecture has very few constraints, it may be arbitrarily deep, nonlinear, recurrent or include attention, covering both toy models and realistic agents. The network learns via gradient updates in the continuous time limit, a regime called gradient flow \cite{saxe_mathematical_2019, boursier_gradient_2022, miyagawa_toward_2023}:
\begin{equation}\label{eq:lr_dynamics}
    \frac{d \mathbf{w}}{dt} = \mu(t) \frac{dP(\mathbf{w}(t))}{d\mathbf{w}}
\end{equation}
where $\mu(t)$ is the agent's learning rate at time $t$. The only condition on the network is that $P(\mathbf w)$ is continuously differentiable in $\mathbf w$ and its gradient $\frac{dP(\mathbf{w}(t))}{d\mathbf{w}}$ is Lipschitz in $\mathbf w$ for all $t$. In practice, this means that all activation functions in our model must be smooth and have bounded first and second derivatives, excluding ReLU-like activations. The agent can influence its learning, i.e. learn faster, slower or quit learning altogether, only by adjusting its learning rate $\mu(t)$, while other aspects of its dynamics, such as the gradient flow learning algorithm, remain fixed.

The \emph{external reward} received by the agent scales with task performance $P(\mathbf{w}(t))$, which generally increases over the episode as the task is learned. At the same time, the agent incurs a \emph{cost} $C(\mu(t))$ for learning at a speed $\mu(t)$. The \emph{internal reward} rate is defined as the difference between these two:
\begin{equation}
    r(t) = P(\mathbf{w}(t)) - C(\mu(t)) \label{eq:internal_reward}
\end{equation}
Balancing effort and performance therefore corresponds to maximizing the discounted integral of the internal reward:

\begin{equation} \label{eq:optimization_goal}
    \mu^*(t) =\underset{\mu(t)}{\arg \max}\int_{t}^{T} d\tau \;\gamma^{\tau-t}\;\underbrace{\left( P(\mathbf{w}(\tau)) - C(\mu(\tau))\right)}_{\textup{internal reward rate}\;  r(t)}
\end{equation}
where $\gamma \in [0, 1]$ is a discount factor reflecting how strongly the agent values future rewards, and $T$ denotes the episode duration.

While the performance function $P(\mathbf{w})$ can take any continuous form, analytical tractability depends on the choice of the cost. An exact solution exists for any function $C(\mu)$ that is convex for $\mu > 0$, and analytical solutions exist whenever the problem $\mu \frac{d C}{d\mu} - C(\mu) = \text{const}$, allows a closed-form expression for $\mu$ (Eq.~\ref{eq:cost_function_requirement} in Methods). In this work, we adopt a quadratic cost in the learning rate where $\beta$ sets the cost of learning: 

\begin{equation} \label{eq:cost_no_alpha}
C(\mu) = \beta \mu^2.
\end{equation}
This setup gives us a simple but general way of thinking about learning as a control problem \cite{bertsekas_dynamic_2000, boscain_introduction_2021, chacko_optimizing_2024}, and similar settings have been used in multiple instances in the cognitive control literature \cite{shenhav_expected_2013, masis_value_2021, piray_linear_2021, agrawal_temporal_2022}. Importantly, in the case of optimizing the learning rate, we derive an analytical expression for closed-loop learning rate scheduling (\autoref{sec:closed_loop}) and, in simpler settings, an open-loop expression in terms of task and agent parameters (\autoref{sec:solution_perceptron}). This contrasts with previous work, in which most solutions are obtained numerically \cite{franceschi_bilevel_2018, carrasco-davis_meta-learning_2024, mori_optimal_2025, mignacco_statistical_2025}.
Our formulation can be applied to arbitrary tasks by appropriately defining the performance measure. For example, in two-alternative forced-choice tasks \cite{bogacz_physics_2006, masis_strategically_2023}, performance may correspond to accuracy in predicting stimulus–reward associations; in next-token prediction tasks, $P(t)$ could be the negative of the cross-entropy loss; and in unsupervised learning problems, $P(t)$ could represent the negative reconstruction error or reconstruction accuracy. Irrespective of the task or model architecture, the same question remains: how should an agent schedule its learning rate $\mu(t)$ so that it can collect high rewards through learning without overspending on effort? In other words, when is it worth learning fast, and when is it better to slow down?

\section{Results}

Determining the optimal learning speed schedule for arbitrary networks and tasks at first glance appears to require full knowledge of the learning trajectory, something analytically tractable only in simple models. The main result of this section is that, without discounted rewards ($\gamma = 1$),  we can bypass this obstacle. We derive a simple analytical expression for the optimal learning rate $\mu^*(t)$ that maximizes the cumulative discounted internal reward during a learning period (Eq.~\ref{eq:optimization_goal}), balancing performance and effort without requiring knowledge of the full trajectory (\autoref{sec:closed_loop}). Remarkably, we show that this solution applies regardless of the task and network architecture, under mild continuity assumptions on learning dynamics. The expression depends on variables the agent might have access to in a naturalistic learning setting, such as its current performance $P(t)$ and its expected final performance $P(T)$. While $P(T)$ may not always be directly available to the learner, we show later how it can be estimated using an episodic memory module to estimate future performance (\autoref{sec:estimating_final_performance}). In special cases where learning dynamics can be solved explicitly, like the perceptron, we show how this general closed-loop expression also leads to a full open-loop schedule (\autoref{sec:perceptron_openloop}). We further explore how task parameters affect the learning rate signal (\autoref{sec:discount_factor_effects}), and the consequences of following suboptimal learning strategies (\autoref{sec:closed_loop}, \autoref{sec:estimating_final_performance}). Finally, we compare the theory to existing experiments (\autoref{sec:animal_behaviour}), and make more predictions in the discussion section.

\subsection{General solution for the optimal learning rate schedule}
\label{sec:closed_loop}
The problem can be framed in standard control-theoretic terms, where the learning rate is the control signal. In continuous-time control problems, the Hamilton–Jacobi–Bellman (HJB) equation \cite{bellman_dynamic_1958, bertsekas_dynamic_2000}, a partial nonlinear differential equation, provides conditions for the optimal value function and the associated control signal. A condition that allows for exact analytical solutions of this partial differential equation is setting the discount factor parameter $\gamma = 1$, which corresponds to an agent that values all future rewards equally. This gives us access to the optimal learning rate schedule (see derivation in the Methods section \autoref{sec:hjb}).
The optimal solution for the learning rate given the learning rule (Eq.~\ref{eq:lr_dynamics}) and the cost of control $C(\mu)$ (Eq.~\ref{eq:cost_no_alpha}) is:

\begin{figure*}
    \centering
    \includegraphics[width=\linewidth]{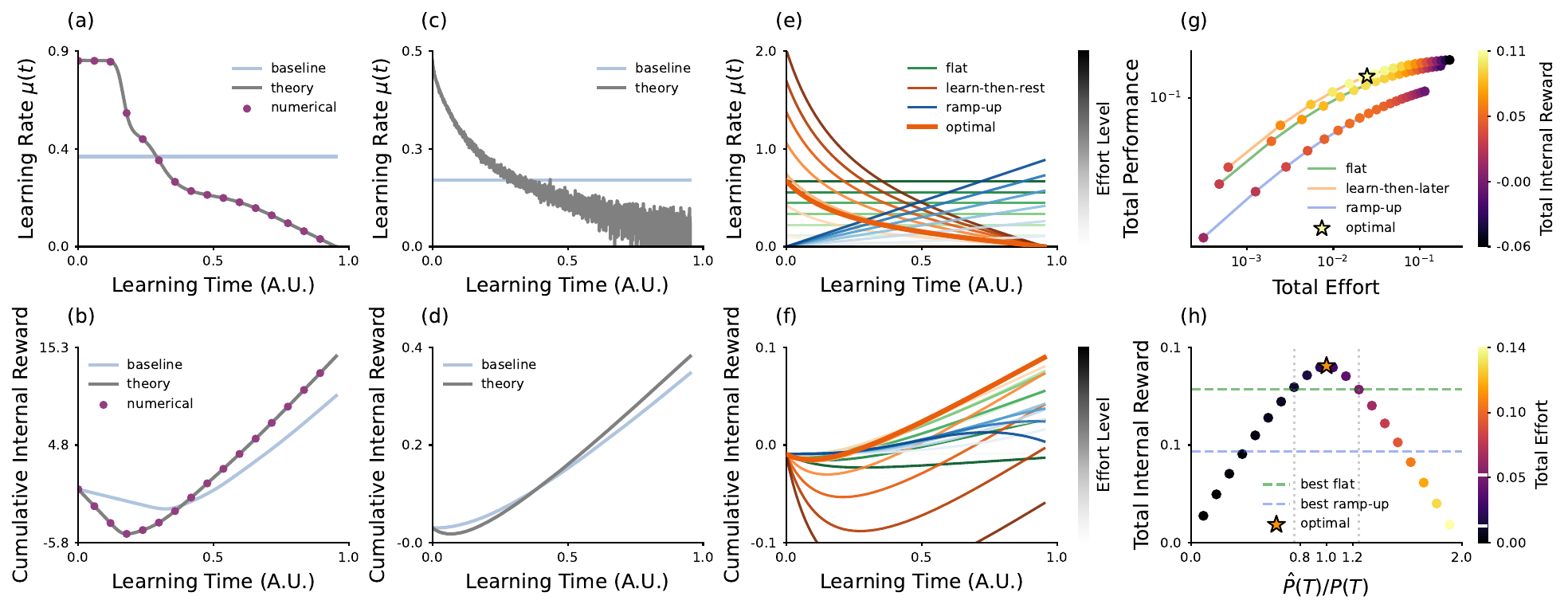}
    \caption{\textbf{Closed-loop expression generalizes to arbitrary tasks and networks.} \textbf{(a,c)}: Optimal learning rate schedules for a two-layer non-linear network on a teacher-student setup trained with gradient flow \textbf{(a)}, and digit classification/MNIST task trained with SGD \textbf{(c)}. \textbf{(b, d)}: Cumulative internal rewards for the optimal solution from \ref{eq:closed_loop} (grey), baseline agent with the best performing fixed learning rate (solid blue), and numerical optimization (purple dots). \textbf{(e, f)} Learning rates (e) and cumulative internal rewards (f) for different learning strategies. Darker tones represent strategies with more effort, colours represent strategy profiles (purple - flat rate, orange - learn-then-rest, blue - ramp-up, bold orange - optimal).
    \textbf{(g)} Plot of performance/cost trade-off for different learning strategies (each dot is a different learning rate schedule). The colourbar shows the cumulative internal reward rate, i.e. the optimization objective. The yellow star shows the optimal strategy. Strategies lie on the lines of learning profiles (colours as in (e,f)). \textbf{(h)} Cumulative rewards of the optimal learning strategy as a function of relative estimation error of final performance $\hat{P}(T)/P(T)$, spanning under- and over-confident regimes (colorbar shows total effort). Coloured dashed lines give the best-performing strategy for each learning profile, with the true optimum indicated by a star. Optimal profiles with estimation errors up to 20\% exceed the best flat strategy (indicated by the vertical gray dashed lines). The same error bounds are marked by white lines on the effort colorbar. Experimental details are provided in section \ref{sec:implementation_details}.}
    \label{fig:closed_loop}
\end{figure*}

\begin{equation}\label{eq:closed_loop}
    \mu^*(t) = \sqrt{\frac{1}{\beta} (P(T) - P(t))}
\end{equation}
where $\beta$ is the cost of effort, $P(t)$ is the performance at time $t$, and $P(T)$ is the performance achieved at the end of the learning episode \emph{under the optimal learning rate schedule $\mu^*$}.

The optimal strategy implied by this solution is intuitive: start by learning quickly and investing substantial effort early on, then gradually reduce effort while continuing to reap the benefits of the earlier gains. This follows directly from the fact that $P(T) - P(t)$ decreases over time because $P(t)$ increases monotonically. When performance is rising quickly, the optimal learning speed should fall just as quickly. Interestingly, if performance plateaus, the prediction is not to increase effort but to maintain the same learning rate and wait it out. Finally, once the agent masters the task and $P(T) - P(t) \to 0$, the optimal strategy is to stop learning altogether. The cost parameter $\beta$ modulates the performance/effort trade-off, such that higher costs lead to lower optimal learning rates throughout.

To exemplify this behaviour, we trained deep non-linear networks on different tasks and set their learning rate according to the optimal learning rate (Eq.~\ref{eq:closed_loop}). One network was trained in a teacher--student setup \cite{goldt_dynamics_2019, lee_maslows_2022}, and another was trained on the MNIST image classification task \cite{deng_mnist_2012}. The teacher--student network was trained in the gradient flow limit and we show that the theoretically optimal learning rate (Eq.~\ref{eq:closed_loop}) matches the numerically optimized learning rate (Fig.~\ref{fig:closed_loop}a,b), following the behaviour previously described. Although the optimal learning rate schedule was derived for the idealized case of full-batch gradient flow learning dynamics, realistic agents, biological or artificial, often learn with batch updates or noisy gradients. The network trained on image classification was updated using stochastic gradient descent (Fig.~\ref{fig:closed_loop}c,d). The theoretical learning-rate schedule is noisy, reflecting fluctuations in the performance signal, yet it consistently outperforms a fixed baseline strategy. This indicates that the theoretical framework remains effective even in noisy learning regimes, where learning must be regulated online from imperfect feedback (see Methods \autoref{sec:implementation_details} for details on network training).

The corresponding cumulative internal reward signal (Fig. \ref{fig:closed_loop}b,d) highlights the effects of the learning strategy: the agent initially pays more in cost than it gains in reward, but this front-loaded effort pays off as performance rapidly increases, leading to higher reward rates that can compensate for the cost incurred in the early effort to speed up learning. In comparison, a baseline agent with a flat small learning rate eventually learns the task, but accumulates substantially less internal reward overall, either underperforming or overspending effort relative to the optimal strategy.

To illustrate how different learning strategies unfold, we compare three qualitatively distinct learning profiles that differ in how effort is allocated over time. The first is a flat learning profile, where a fixed learning rate and constant effort are maintained throughout the learning episode. The second is a ramp-up profile, where the agent starts slowly and gradually increases effort. The third is the optimal “learn fast, rest later” profile derived from our theory. For each profile, we vary the maximum level of effort to create different learning strategies (Fig.~\ref{fig:closed_loop}e) and then we compare their performance, costs, and overall objective (total internal reward). The cumulative internal reward rate curves for the three strategies differ markedly, with the optimal strategy resulting in the highest total internal reward (Fig.~\ref{fig:closed_loop}f). The trade-offs between performance and effort are likewise revealed by the network learning: as expected, higher effort leads to greater external reward, but with diminishing returns beyond a certain point (Fig.~\ref{fig:closed_loop}g). Interestingly, while multiple strategies can yield similar performance–effort ratios (for instance, the flat and optimal schedules), only the optimal shape with the appropriate effort level maximizes the cumulative internal reward. Moreover, getting the form of the learning strategy right matters more than fine-tuning the exact effort level—an agent following the right qualitative schedule remains close to optimal even with moderate misestimates of effort (Fig.~\ref{fig:closed_loop}h).

The closed-loop solution for the optimal learning rate (Eq.~\ref{eq:closed_loop}) provides a normative benchmark: it tells us what the optimal learning rate would be if the agent had access to both its current performance $P(t)$ and its final performance $P(T)$. In cases where learning leads to an improvement in performance, the optimal learning rate profile exhibits a "learn fast, rest later" shape: starts with high learning rate, and as performance approaches its asymptotic value, the learning rate is decreased. To the best of our knowledge, we provide a novel closed-loop analytical solution to the learning rate control problem, which can be applied to other tasks and architectures as previously discussed, whereas prior efforts primarily rely on numerical optimization to compute the optimal learning rate scheduling. In practice, while $P(t)$ is observable most of the time (except in cases with sparse reward or noisy feedback), $P(T)$ is not, as it requires knowledge about the future time evolution of the learning system. This raises a natural question: how could a real agent approximate such a quantity, and what happens when the estimate is inaccurate? As previously discussed, minor to moderate estimation errors have little impact, but large misestimations of $P(T)$ lead to significantly worse outcomes than even the simplest flat learning strategies (Fig.~\ref{fig:closed_loop}h). These considerations motivate the next section, where we propose a mechanism to estimate $P(T)$ based on episodic memory, and examine the consequences of misestimation on learning performance and effort.

\begin{figure*}
    \centering
    \includegraphics[width=\linewidth]{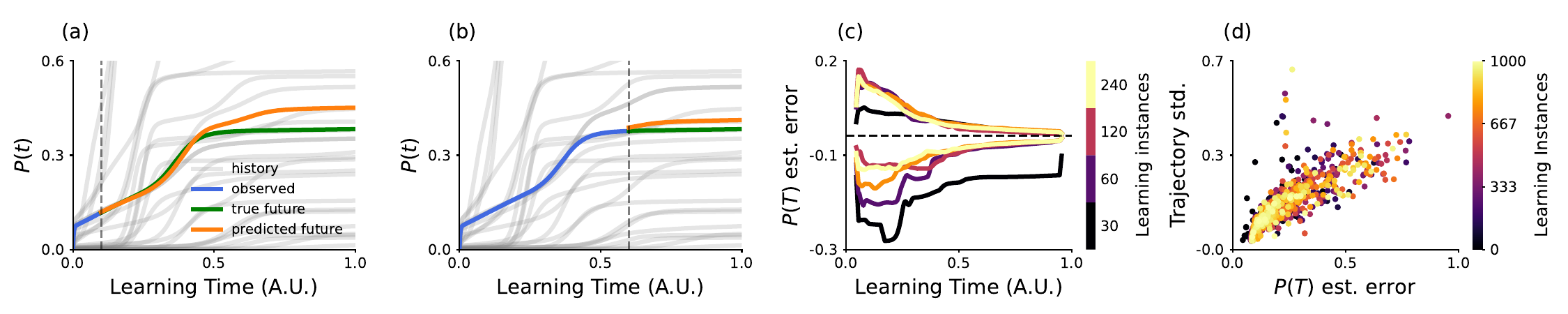}
    \caption{\textbf{Estimating final performance through episodic memory.} 
\textbf{(a, b)}: Estimated future trajectory (orange) throughout learning, based on the observed trajectory $\mathcal{T}_{o}$ (blue), using similarities with trajectories stored in memory $\mathcal{T}_{h}$ (gray). True future trajectory shown in green. 
\textbf{(c)}: Estimation error $\hat{P}(T|t)-P(T)$  as a function of within-episode time 
$t$. For each memory size (colour; number of past learning episodes used), we plot the 25th and 75th percentiles of the error across trials (smoothed). As memory increases and more of the current trajectory is observed, the error both contracts and becomes increasingly symmetric, indicating a reduction in bias as well as variance.
\textbf{(d)}: Learning-trajectory variability (measured as standard deviation) versus estimation error, coloured by the order of learning instance in memory. Higher-variance trajectories exhibit larger estimation errors. Later learning instances cluster near low variability and near-zero error, yet even at late stages increased trajectory variability remains associated with reduced predictability.
}
    \label{fig:PT_estimation}
\end{figure*}

\subsection{Estimating final performance through episodic memory} \label{sec:estimating_final_performance}
According to the proposed theory, optimal learning requires accurate estimation of task performance at the end of the episode, $P(T)$, under the optimal learning rate schedule, $\mu^*(\cdot)$. We hypothesize that estimating this quantity constitutes a form of meta-learning, in which an agent learns to predict how its own improvement will evolve based on prior learning episodes.

We propose a simple yet biologically plausible model for estimating future performance based on episodic memory. Each stored memory corresponds to a past learning trajectory, $\mathcal{T}(t)$, defined by performance over time, $\mathcal{T}(t) = \{P(0), P(dt), \ldots, P(t)\}$. When facing a new task, the agent compares its ongoing trajectory, $\mathcal{T}_o(t)$, with $H$ previously stored trajectories $\{\mathcal{T}_h(t)\}_{h\in H}$ retrieved from memory. Using similarity-based weights $\{w_h(t)\}_{h \in H}$ computed with a Gaussian kernel $w_h(t) \propto k\left(\mathcal{T}_o(t), \mathcal{T}_h(t)\right)$ up to the current time $t$, the agent extrapolates its future performance at $T > t$:
\begin{equation}
    \hat{P}(T \mid t) 
= \sum_{h \in H} w_h(t) \, P_h(T) 
\end{equation}
At the beginning of each task ($t = 0$), when no trajectory has yet been observed, all previous experiences contribute equally to the initial estimate $\hat{P}(T \mid t=0)$.

We evaluate this estimator across 1,000 sequential learning episodes. Each episode involves a nonlinear teacher network with randomly sampled weights, while the agent network is randomly initialized each time and trained on teacher-generated data, producing diverse learning dynamics. As each episode progresses, the agent continuously refines $\hat{P}(T \mid t)$ based on the observed trajectory (Figure~\ref{fig:PT_estimation}a,b) and uses it to compute the evolving optimal learning rate $\mu(t)$ (see Methods~\ref{sec:estimation_PT_supp}). Once the time limit $t=T$ is reached, the complete trajectory $\mathcal{T}(T)$ is stored in memory before the next learning episode begins.

As more of the current trajectory is observed, the similarity computation relies on longer segments of the trajectory, improving predictions of the remaining performance (Fig.~\ref{fig:PT_estimation}c). The estimation accuracy becomes less variable and less biased as more learning trajectories accumulate in memory (Fig.~\ref{fig:PT_estimation}c), enabling better reconstruction of the learning dynamics and lower deviations of $\mu(t)$ from the theoretical optimum $\mu^*(t)$. This demonstrates a form of meta-learning, where better control of learning is achieved as more learning episodes are experienced. However, estimation accuracy remains strongly modulated by the intrinsic variability of the learning dynamics: trajectories with higher variability are systematically harder to predict and incur larger estimation errors, even at later learning stages (Fig.~\ref{fig:PT_estimation}d).

Although performance estimation via episodic memory depends on the structure of the task distribution, these results show that a simple memory-based mechanism that compares the current trajectory with previously experienced ones can achieve near-optimal behaviour. This provides a plausible route for both biological and artificial agents to meta-learn, throughout their life span, how to allocate learning effort when encountering unseen tasks.

\subsection{Mathematical analysis for a linear perceptron}\label{sec:solution_perceptron}

\begin{figure*}[t]
    \centering
     \includegraphics[width=\linewidth]{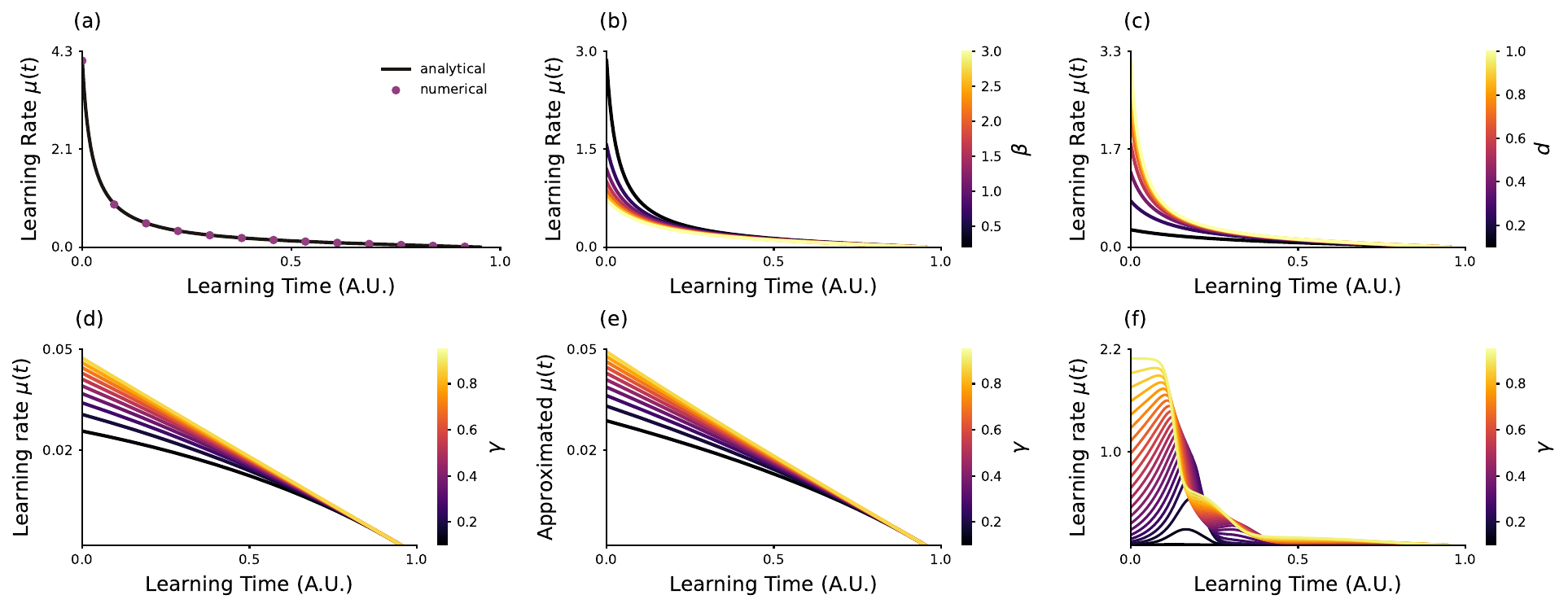}
    \caption{\textbf{Optimal learning strategies change with parameters in predictable ways.} \textbf{(a-c)} A linear perceptron learning a regression task. \textbf{(a)} Optimal learning rate (dots: numerically optimized, line: analytical expression from Eq. \ref{eq:lr_openloop_no_alpha}). \textbf{(b)} Increasing the cost of control coefficient $\beta$ leads to lower learning rates in a non-linear way. \textbf{(c)} Task difficulty, defined as the norm of the difference between the target and initial weights $d = \| \mathbf w^* - \mathbf w_0\|^2$, increases optimal learning rate. \textbf{(d,e)} Learning rate for a one linear neuron learning a linear regression task with discounted time, an analytical approximation from Eq. \ref{eq:gamma_approximation} (e), or numerically evaluated (d). The approximation retains qualitative properties of the exact solution, showing that stronger discounting leads to lower learning rates. \textbf{(f)} Two-layer linear network learning a teacher-student regression task, showing a non-linear effect of changing $\gamma$ on the optimal learning rate. Overall, decreasing the cost of effort, increasing task difficulty or increasing the discount factor scales the optimal learning rate up. Furthermore, strong discounting can completely alter the shape of learning schedule when nonlinear learning dynamics is present. Experimental details are provided in section \ref{sec:implementation_details}.}
    \label{fig:open_loop}
\end{figure*}

While the closed-loop solution prescribes the optimal learning rate online (Eq.~\ref{eq:closed_loop}), it offers limited insight into the resulting training dynamics. Here, we derive an open-loop solution that characterizes the full optimal learning trajectory. The closed-loop solution is ideal for agents that need to adapt in real time, but it makes it difficult to analyse the full trajectory in detail. For example, in the previous nonlinear network training examples (Fig.~\ref{fig:closed_loop}a), the learning rate shows drops and plateaus that mirror the shape of the performance curve. We might anticipate such plateaus in deep nonlinear networks, but their timing cannot be predicted in advance. Similarly, for architectures like RNNs or transformers, the trajectory of $\mu(t)$ can only be analysed after training. In contrast, for simple models such as a linear perceptron, the learning dynamics are known explicitly under some assumptions on the task, data distribution and output non-linearities \cite{horner_dynamics_1993, saad_-line_1995}, which allows us to substitute them into the closed-loop expression and derive a full open-loop solution.

Consider a linear perceptron with weights $\mathbf{w}$ trained on a linear regression task, with learning rate costs (Eq.~\ref{eq:cost_no_alpha}) and external rewards defined as the negative mean squared error (MSE) loss, $P(t) = -L(t)$ ($\eta=1$ in Eq.~\ref{eq:internal_reward}). Training inputs are drawn i.i.d.~from $\mathbf{x} \sim \mathcal{N}(0, \sigma^2 \mathbf{I})$, with labels generated by target weights $y = \mathbf{x}^\text{T}\mathbf{w}^*$. The network is initialized with weights $\mathbf{w}_0$, and the difficulty of the task is captured by the squared distance $d^2 = \|\mathbf{w}^* - \mathbf{w}_0 \|^2$.

Using this setup to compute the optimal learning rate (Eq.~\ref{eq:closed_loop}) yields a self-consistent integro-differential equation for $\mu^*(t)$ (see Methods section \ref{sec:perceptron_openloop}). Solving this equation gives the following open-loop expression:
\begin{align}
 \mu(t) =  \theta \tan\left(\sigma^2 \theta (T-t) \right)& \label{eq:lr_openloop_no_alpha}\\
    \theta = \frac{\sigma d}{\sqrt{2\beta}} \cos \left(\sigma^2 \theta T \right)& \label{eq:theta}
\end{align}
where $\theta$ is obtained numerically from the self-consistent Eq.~\ref{eq:theta}.

The open-loop form of the solution preserves the intuition of the closed-loop form: since $\tan(T-t)$ is a decreasing function of time, the optimal schedule begins with rapid learning and gradually slows down. We verify that the open-loop expression matches the numerically optimized learning schedule in Fig.~\ref{fig:open_loop}a. This explicit solution allows examining how the optimal schedule depends on task parameters. While the qualitative shape of the curve is stable, lower effort cost and greater task difficulty justify higher learning rates (Fig.~\ref{fig:open_loop}b,c).

So far, we have assumed that agents value all future rewards equally ($\gamma = 1$). For tasks on very long timescales, this assumption may be unrealistic, and it becomes important to consider how optimal learning strategies change once future rewards are discounted.

\subsection{Time discount effects} \label{sec:discount_factor_effects}

Incorporating a discount factor $\gamma~<~1$ substantially changes the control problem by prioritizing near-term improvements over rewards that lie further into the future. Discounting captures the agent's uncertainty about future outcomes, can be used to account for bounded computational resources for long-horizon predictions, and ensure the stability of learning in stochastic environments. In contrast to the undiscounted case, where learning effort can be allocated across the entire learning episode, the agent now prioritizes outcomes in the near future, placing progressively less weight on rewards further ahead.

The presence of discounting breaks analytical tractability of the Hamilton-Jacobi-Bellman equation. Nevertheless, approximate analytical solutions can still be obtained for linear models using the Homotopy Perturbation Method with Padé approximants \cite{ganjefar_modified_2016}, a method well established in control theory literature \cite{barari_application_2008, abdulaziz_application_2008, buhe_application_2023} (see Methods section \ref{sec:homotopy_pade}). This approach produces a sequence of increasingly accurate analytical approximations that, under some parameter regimes, converge toward the true solution, with the first few modes already capturing the dominant qualitative effects of the discount factor $\gamma$.

We first consider a single neuron trained on a linear regression task with mean squared error loss. The leading order approximation provides a simple and interpretable expression for the optimal learning rate schedule:

\begin{equation}\label{eq:gamma_approximation}
    \mu^*(t) \approx \frac{1}{2\beta} \frac{\left(\frac{dP}{d w}\right)^2}{\frac{1}{T-t} - \frac{1}{2}\log \gamma}
\end{equation}
The approximation makes explicit how discounting suppresses learning rate magnitude: decreasing the discount factor reduces the overall effort allocation and lowers the learning rate ($\log \gamma$ term). The approximation also predicts the reduction of control near the end of the learning episode ($T-t$ term) and when performance gradients saturate ($\left(\frac{dP}{d w}\right)^2$~term). Despite its simplicity, this leading order solution closely matches the numerically optimized learning rate schedule (Fig.~\ref{fig:open_loop}d,e).

For more complex architectures, analytical approximation become increasingly restrictive, and we therefore analyze numerically optimized solutions. In a two-layer linear network (Fig.~\ref{fig:open_loop}f), the undiscounted solution ($\gamma = 1$) displays prominent drops and plateaus reflecting the stage-wise learning dynamics induced by weight coupling \cite{saxe_mathematical_2019, braun_exact_2022}. As the discount factor is reduced, these features gradually fade as the model becomes increasingly short-term oriented ($\gamma \ll 1$).

Interestingly, for sufficiently small $\gamma$ values, the optimal learning rate schedule enters a qualitatively distinct regime (see Fig.~\ref{fig:open_loop}f). Learning initially remains near zero,  followed by a gradual ramp-up and subsequent decrease in the learning rate. This behaviour reflects a rational allocation of effort under strong discounting and nonlinear learning dynamics. Early learning is relatively expensive because gradients are low and little time remains to convert effort into performance gains, thus the optimal strategy conserves learning rate effort initially and allocates it later, when the learning gradients are larger and improvements per unit cost are higher. This regime differs from the single-neuron case and arises from the interaction between discounting and the stage-like learning dynamics characteristic of deeper networks.

Together, the analytical approximations as well as the closed- and open-loop solutions show that larger discount factors, lower effort costs, and greater task difficulty license larger learning-rate schedules. In contrast, when future rewards are strongly discounted, learning effort becomes progressively less worthwhile, leading to slower or delayed learning even when improvements remain possible.

\subsection{Connection to animal behaviour} \label{sec:animal_behaviour}

\begin{figure}[t]
    \centering
     \includegraphics[width=0.6\linewidth]{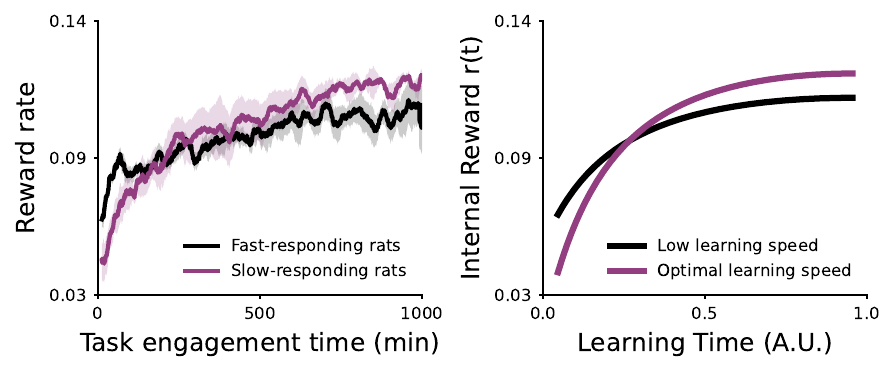}
    \caption{\textbf{Comparison to data.} Left: Instantaneous reward rate over time in a two-alternative forced choice task. Rats were grouped by reaction times relative to their individual baselines from prior experiments (fast, black; slow, purple). Data adapted from \cite{masis_strategically_2023}, Fig. 7b. Right: Neural network model performing a two-Gaussian discrimination task. One model follows the optimal learning strategy (purple), while the other uses the same profile with a reduced learning-rate magnitude (black). Animals exhibit a trade-off between immediate reward rate and learning speed: lower immediate rewards, for example due to longer reaction times under greater control, can yield faster learning and higher cumulative reward. Our framework captures this trade-off as a balance between control cost and learning rate. Experimental details are provided in section \ref{sec:implementation_details}.}
    \label{fig:rat_data}
\end{figure}

We next examine how the predictions of our framework align with animal behaviour during learning. While, to the best of our knowledge, there are no experiments that systematically test our theory of learning speed control and effort costs, there are notable similarities between our framework and ideas from the cognitive control literature. In particular, an experiment by Masis et al. \cite{masis_strategically_2023} aimed to test whether long-term learning is taken into account in animal decision-making tasks, and showed that there is a trade-off between collecting short term rewards and learning speed. Rats were trained on a binary classification task, in which an image was presented on both the left and right sides of their field of view, and only one of the images yielded a reward, indicating their choice by licking a left or right port (Fig.~5 in \cite{masis_strategically_2023}). Rats were grouped according to their reaction times at the start of the learning episode. Those with longer reaction times collected less reward per unit of time but learned faster compared to rats with shorter reaction times at the start of training. Moreover, slower-responding rats reached higher instantaneous rewards sooner than the fast-responding rats (Fig.~\ref{fig:rat_data}a). These results are recapitulated, with larger effect sizes, when rats are experimentally constrained to respond either fast or slow, as opposed to responding freely. (\cite{masis_strategically_2023}, Fig. 5).

We compare this result with our framework and show that two artificial agents with lower and higher amount of learning rate produce qualitatively similar reward rate curves to the two groups of rats in the experiment. In this comparison, we use internal reward rates from our model and the instantaneous reward rates from the behavioural data, noting that longer response times could effectively correspond to higher costs of effort, i.e. the cost of speeding up learning in these rats is the opportunity cost caused by a reduction in a response rate. 
We cannot make claims about what are the discount factors or costs of effort for the two groups of rats, or exclude the possibility that rats may have been using a slightly different learning strategy, which could yield similar reward rate curves. With experiments designed more precisely to test the problem we consider here, such hypotheses could be directly evaluated.

What we highlight with this animal experiment is the existence of an intertemporal choice in the control allocated to learning, where effort and its associated cost can impact the resulting learning trajectory and, consequently, the cumulative reward. Our framework formally addresses this fundamental problem, and we later argue and discuss that such choices are ubiquitously present in animal learning, especially in humans, where long-term planning and lifelong learning lie at the core of developing higher level cognitive abilities.

\section{Discussion} \label{sec:discussion}

In this work, we formalized the problem of scheduling learning speed as a normative optimization balancing effort and performance, specifically by optimizing a total internal reward throughout learning. The internal reward rate increases with task performance, while exerting more effort decreases it (\autoref{sec:the_framework}). This framework applies to learning in both artificial and biological agents, as it can be used to inspect learning of various model architectures, spanning simple toy models and realistic non-linear networks, and applies beyond the tasks presented in this paper due to its abstract formulation. Additionally, our framework is mathematically tractable, providing a closed-loop solution (\autoref{sec:closed_loop}) that depends on quantities estimable across multiple learning episodes (\autoref{sec:estimating_final_performance}). We can fully characterize the learning trajectory and obtain open-loop solutions for simpler models (\autoref{sec:perceptron_openloop}) and approximations for time-discounted rewards (\autoref{sec:discount_factor_effects}). This provides access to explicit mathematical expressions showing that the learning rate increases with lower effort costs, increased discount factor, and greater task difficulty. This allows us to understand control over learning from first principles, through the normative objective of maximizing reward.

Below, we expand the discussion of our work by examining its implications for animal behaviour (\autoref{sec:animal_behaviour}) and drawing connections to the psychology literature, specifically on self-regulated learning \cite{bandura_self-efficacy_1977, zimmerman_attaining_2000} and cognitive effort in the context of learning \cite{shenhav_expected_2013, kool_mental_2018, masis_strategically_2023, masis_learning_2024}. We then explore links to neuroscience, as many of the quantities required (according to our theory) to compute optimal learning effort correlate with neural activity and neurotransmitter dynamics in the brain \cite{doya_metalearning_2002, aston-jones_integrative_2005, shenhav_expected_2013}. Finally, while our normative objective aims to account for natural behaviour, our framework also has implications for machine learning, as it specifies a way to optimize learning. Learning rate schedules have been studied extensively in machine learning \cite{choi_empirical_2020, schmidt_descending_2021, kashyap_survey_2023} and is particularly important given the widespread use of large architectures.

\subsection*{Performance estimation as self-efficacy}
The optimal learning rate schedule (Eq.~\ref{eq:closed_loop}) suggests allocating effort in proportion to the perceived learning gains, quantified as the difference between expected future and current performance, $P(T) - P(t)$. This idea aligns closely with the literature on self-regulated learning \cite{bandura_self-efficacy_1977, pajares_role_1994, zimmerman_attaining_2000, zimmerman_self-regulated_2011}, which proposes that \textit{self-efficacy}, the learner's belief in their capacity to successfully perform a learning task, affects how much effort they invest in the task and how persistent they are.

Self-efficacy differs from simple action-outcome expectations. First, because individuals may understand what behaviour leads to success but still doubt their capacity to perform it. Second, it requires some level of meta-cognition of their own capacity to learn, and the effect of effort in their own learning. In our formulation, this belief is captured by the perceived gap between what one can currently achieve and what one believes they can achieve, $P(T) - P(t)$, which determines the optimal level of learning effort. Consistent with this, self-efficacy predicts performance in problem-solving tasks more strongly than the objective difficulty of the problems \cite{pajares_role_1994, kontas_explaining_2022}, paralleling our prediction that inaccurate estimates of $P(T)$ lead to suboptimal learning speeds and poorer outcomes regardless of task difficulty (Sections \autoref{sec:closed_loop} and \autoref{sec:estimating_final_performance}).

Learning to accurately estimate one’s own future performance thus connects directly to the development of self-beliefs and self-regulatory skills, both of which are essential for adaptive learning and long-term success \cite{zimmerman_self-regulated_2011}. Along these lines, a limitation of our framework is that it considers the optimization of the learning rate for a single task, whereas in principle the control signal could be multidimensional. In more complex settings, an agent might need to maximize cumulative reward across multiple concurrent tasks, by distributing its learning effort to each one, thereby deciding not only \textit{how much} to learn but also \textit{what} to learn. Related analyses have been explored in the past in simplified learning settings \cite{son_metacognitive_2006, son_metacognitive_2010}. Extending our theory in this direction could provide a principled account of how learning effort is allocated to multiple tasks under limited attentional resources.

\subsection*{Learning effort and task difficulty} The optimal learning rate expression also links conceptually to the cost of learning effort. It predicts that less effort should be invested when learning is more costly (larger $\beta$ implies smaller $\mu$, Eq. \ref{eq:closed_loop}), effectively treating the learning rate schedule as a schedule of effort over time. For extremely difficult tasks (large gap between initial performance $P(0)$ and final performance $P(T)$), the theory predicts very high initial effort, which may not be realistic for biological learners. It is plausible that individuals have an internal effort threshold that they are unwilling or unable to exceed. If mastering a task requires levels of effort $\beta\mu^*(t)^2$ that surpass this threshold, learners are expected to disengage or give up on the optimal strategy, as it has been observed in non-learning behavioural experiments \cite{kool_decision_2010, kool_mental_2018}. This idea resonates with both cognitive-load-based accounts of effective curricula \cite{sweller_cognitive_1988, van_merrienboer_cognitive_2005} and findings from the self-regulated learning literature \cite{bandura_cultivating_1981, schunk_goal_1990}. By breaking a complex task into smaller, more manageable components, an approach known as \textit{proximal subgoals} in the self-regulated learning literature, learners can improve self-efficacy, sustain progress and reduce the likelihood of abandoning learning prematurely \cite{stock_proximal_1990}.

Conversely, abrupt increases in task difficulty can disrupt learning. Empirical studies show that students often experience a decline in academic performance at key educational transitions (e.g. from middle to high school, or high school to university) \cite{rice_explaining_2001, neild_falling_2009}. These effects are partly social \cite{evans_review_2018}, partly due to changes in academic demands \cite{hills_transfer_1965}, and partly psychological \cite{lin_students_2023}. Within our framework, such phenomena could arise when learners misestimate their performance trajectory. For instance, students who previously excelled may enter a new environment overconfidently, overestimating their current performance (high $P(t)$) and thus perceiving a smaller $P(T) - P(t)$ gap than the true value. This would lead them to invest too little effort into learning and be surprised by low test scores. Conversely, students who underestimate their current ability but overestimate future performance (low $P(t)$, high $P(T)$) may exert excessive effort early on, and if performance improvements fail to meet expectations, the resulting mismatch between effort and reward could contribute to burnout \cite{salmela-aro_role_2008, barbosa_transition_2016}. While these correspondences are only qualitative, these patterns illustrate how errors in estimating $P(T)$ or in exceeding effort constraints could link to real-world learning trajectories.

\subsection*{Online effort regulation} Our framework also suggests how agents could, in principle, compute the optimal learning speed on the fly. In the model, current performance $P(t)$ represents an observable measure of task success, while the expected final performance $P(T)$ must be estimated, potentially through a meta-learning process that draws on prior learning experiences. In real-world settings, $P(t)$ is often only sparsely observable (for example, when feedback is limited to intermittent test scores), so both $P(t)$ and $P(T)$ may need to be inferred dynamically using an episodic-memory–like estimator. Ongoing estimates can generate prediction errors, differences between expected and observed performance, that may drive diverse learning behaviors such as disengagement, overconfidence, and burnout as previously discussed. In the self-regulated learning literature, frequent and accurate feedback has been shown to improve engagement \cite{schunk_goal_1990, schunk_learning_1991}, which in our framework corresponds to having precise and timely estimates of $P(t)$ to optimize engagement, while self-efficacy could map to the estimate of final performance $P(T)$.

\subsection*{Effort and neuromodulation} This view connects naturally to known neuromodulatory systems involved in regulating learning and effort. Dopamine encodes reward prediction errors \cite{glimcher_understanding_2011, schultz_dopamine_2016, taira_complementary_2025}, and may signal deviations between expected and observed performance such as $P(t)$ or $P(T)$. Serotonin may support long-timescale regulation and patience \cite{miyazaki_optogenetic_2014, miyazaki_reward_2018, taira_complementary_2025}, and may contribute to estimating $P(T)$ and moderating premature effort allocation. It has also been implicated in overcoming effortful actions by reducing perceived effort costs \cite{meyniel_specific_2016}, suggesting a role in adjusting the effective cost parameter $\beta$. Noradrenaline (NE) and the activity in locus coeruleus (LC), encodes the cost of effort and mobilizes cognitive control \cite{varazzani_noradrenaline_2015, aston-jones_integrative_2005, poe_locus_2020}, while pupil dilation, an LC–NE proxy, tracks subjective effort \cite{gilzenrat_pupil_2010}. Finally, the dorsal anterior cingulate cortex integrates reward, cost, and time information to guide control allocation \cite{shenhav_expected_2013}, and could thus implement computations similar to those determining the optimal learning rate schedule $\mu^*(t)$. Based on these experimental observations, broader computational roles have been attributed to neuromodulators within a meta-learning or continual learning framework. These neuromodulators are thought to contribute not only to learning itself but also to \textit{learning to learn}, by modulating attention, perceived effort, time discounting, and other variables operating in the outer loop of task learning \cite{doya_metalearning_2002, lee_lifelong_2024}. Our work follows a similar conceptualization, in which learning rate control is optimized for a meta-objective (maximizing the time-integrated reward and improving $P(T)$ estimation across multiple learning episodes) rather than for task-specific instant performance.

We believe that extending the applicability of our framework to other learning systems, such as those relying on rules beyond gradient flow, including Hebbian learning \cite{flesch_modelling_2022}, feedback alignment \cite{lillicrap_what_2019}, or efficient coding principles \cite{song_inferring_2024}, is a promising direction for bringing our abstract formulation closer to biologically plausible architectures. Additionally, cognitive control models such as the Stroop network \cite{cohen_control_1990, musslick_rational_2020} apply control to multiple neurons simultaneously, and in some cases the control signals may be highly localized \cite{aston-jones_integrative_2005, shenhav_expected_2013}. As noted earlier, our control signal is currently a scalar, solvable within our specific setup of learning-rate control and gradient flow as the plasticity rule, while in principle still being adaptable to a broad class of learning systems.

Because our framework is abstract, translating it directly to specific experimental setups presents challenges. Assessing whether animals implement processes resembling those proposed here for controlling learning rate would require carefully designed psychological or neuroscience experiments that track the relevant quantities, such as beliefs or estimates of final performance under optimality, continuous monitoring of task performance, and subjective perceptions of effort. An even more open question is how such estimates of final performance might be formed. Episodic memory is one possible mechanism, but alternatives such as model-based estimation or internal models of effort-regulated learning dynamics may also play a role. Another possibility is that learning rate scheduling is carried out in a model-based manner, using a dynamical system that can predict performance and the effect of the learning rate on that system \cite{pinon_model-based_2022, dong_adalrs_2025}.

\subsection*{Scope and limitations}
Our framework optimizes cumulative performance over learning while imposing an explicit cost on the learning rate, in contrast to standard machine learning practice where learning rates are treated as cost-free hyperparameters and performance at convergence is the primary metric \cite{kashyap_survey_2023, choi_empirical_2020, schmidt_descending_2021}. Accordingly, the framework is not intended as a prescription for conventional deep learning optimization, but as a normative account of how learning speed should be regulated when learning itself is costly.

Formally, the problem can be viewed as an optimal control task in which the learning rate modulates the speed of gradient-based learning dynamics, with the solution obtained by solving a Hamilton–Jacobi–Bellman equation. While this structure resembles classical control approaches such as Linear Quadratic Regulator formulations \cite{augustine_note_2023, chacko_optimizing_2024}, those methods rely on linear dynamics and quadratic costs that are violated in learning systems whose dynamics depend nonlinearly on their own performance. Other control-theoretic tools, such as Kalman filtering, optimize instantaneous state-estimation error under noise and target a different objective \cite{piray_model_2021, si_review_2025}.

The framework is also conceptually related to value functions in reinforcement learning and to meta-learning approaches that optimize hyperparameters in an outer loop \cite{shenhav_expected_2013, carrasco-davis_meta-learning_2024}. However, unlike reinforcement learning, where value functions are learned from nonstationary experience, the value function here is computed directly from known learning dynamics; extending this approach to reinforcement learning agents would therefore require tractable descriptions of value function dynamics during policy learning.

Finally, the learning rate is treated as a continuous control variable, whereas biological systems are likely subject to hard constraints not captured by the model. The precise form of the cost associated with exerting control over learning is not known empirically, but making such costs explicit provides a principled way to characterize trade-offs between learning speed, effort, and performance, rather than a mechanistic account of biological implementation.

\subsection*{Closing remarks} Our work consists of several components: a normative framework for optimizing learning, a mathematical derivation of its solution from first principles, simulation-based validation, and connections to theories of cognitive control and self-regulated learning. We argue that formalizing optimal learning rate scheduling through a normative objective offers a promising approach to understanding how animals regulate their own learning over lifespan timescales. The ability to decide when and how much effort to invest in improving one’s performance is central to high-level cognition, shaping motivation, persistence, and adaptability in changing environments. At the same time, our framework offers insights into how such principles might be generalized beyond biological learners. The same mechanisms that allow agents to allocate effort efficiently could inform the design of artificial systems capable of extending their own learning over time, integrating motivation-like control into the dynamics of improvement itself.

\section{Methods}

\subsection{Deriving the closed-loop solution}\label{sec:hjb}

The optimal learning rate $\mu^*(t)$ is the maximizer of the right-hand side (RHS) of Equation \ref{eq:optimization_goal}. It turns out that the expression on the RHS is a well-studied object in control theory and continuous reinforcement learning, known as the optimal value function $V(t=0, \mathbf{w}_0)$ \cite{bellman_dynamic_1958}. This function is defined for any given starting time of the trial $t$ and any initial weight vector  $\mathbf{w}$, and corresponds to the cumulative reward under the optimal control and corresponding weight dynamics starting from $(t, \mathbf{w})$:

\begin{equation}\label{eq:value_fn}
    V(t, \mathbf{w}) = \max_{t \leq \mu(\tau) \leq T} \int_t^T d\tau \;\gamma^{\tau-t}\;r(\tau)
\end{equation}
where the reward rate is $r(\tau) = P(\mathbf{w}(\tau)) - C(\mu(\tau))$. By using the Bellman's optimality principle and manipulating resulting equations, we can derive a Hamilton-Jacobi-Bellman equation, which is a continuous version of the Bellman equation describing the value function \cite{yong_dynamic_1999, bellman_dynamic_1958}:

\begin{equation} \label{eq:hjb}
    \log \gamma \;V(t, \mathbf{w}) + \frac{\partial V}{\partial t} + \max_\mu \left\{ r(t) + \frac{\partial V}{\partial \mathbf{w}}^\textup{T} \frac{d\mathbf{w}}{dt} \right\} = 0
\end{equation}
Let us explain the intuition behind this equation. The first two terms are negative, and roughly correspond to the decrease in the value function due to the discount factor $\log \gamma \; V(t, \mathbf{w})$, and due to the passing of time $\frac{\partial V}{\partial t}$ (even the state doesn't change, the passing time reduces the available value). The second two terms tell us how much reward $r(t)$ one can obtain right now by applying a control $\mu$, and $\frac{\partial V}{\partial \mathbf{w}}^\textup{T} \frac{d\mathbf{w}}{dt}$ tells us how the value will change (for better or worse) by applying this control and updating the state by the resulting $\frac{d \mathbf{w}}{dt}$. The full equation states that for the optimal value function, there is a balance between value lost due to reward discounting and time passing, and value gained due to earned rewards and shifting the state to a more valuable region.

The optimal control choice is hence the control that maximizes the sum of the instantaneous reward and the value change caused by this control. The maximization term is also called the Hamiltonian \cite{yong_dynamic_1999}, which we can further specify by substituting the internal reward rate and learning dynamics expressions from Eqs. \ref{eq:internal_reward} and \ref{eq:lr_dynamics}:

\begin{equation}\label{eq:hamiltonian}
  H(\mu, \mathbf{w}, t)
  = P(\mathbf{w}) - C(\mu) + \mu \frac{\partial V}{\partial \mathbf{w}}^\textup{T} \frac{dP}{d \mathbf{w}} 
\end{equation}

\noindent Maximizing such Hamiltonians results in control expressions that depend only on the value function, but finding the value function involves fully solving the Hamilton-Jacobi-Bellman equation (Eq.~\ref{eq:hjb}). This equation is a nonlinear PDE and it does not have closed form solutions in its general form. Often, the only way of solving it is via numerical methods, or through lengthy analytical approximations \cite{beard_approximate_1998, saberi_nik_approximate-analytical_2012, ganjefar_modified_2016}. However, under certain simplifications, exact analytical solutions can be found, as we show below. 

We can rewrite the Hamiltonian in terms of its maximizer:

\begin{align}
    &\frac{dH}{d\mu} = -\frac{dC}{d\mu} + \frac{\partial V}{\partial \mathbf{w}}^\textup{T} \frac{dP}{d \mathbf{w}}  = 0 \\
    &H(\mu^*, \mathbf w, t) = P(\mathbf w) - C(\mu^*) + \mu^* C'(\mu^*)
\end{align}
where $C'(\cdot)$ denotes its derivative. Substituting back into Eq. \ref{eq:hjb} leads to:
\begin{equation}
    \log \gamma \;V(t, \mathbf{w}) + \frac{\partial V}{\partial t} + P(\mathbf w) - C(\mu^*) + \mu^* C'(\mu^*)= 0
\end{equation}
Then, we note that gradient flow learning problem doesn't intrinsically depend on the time, but only on the initial weights $\mathbf w_0$, the loss and the schedule of the learning rate. This allows a separable solution for the value function:

\begin{align}
    &V(t, \mathbf w(t)) = \Phi(\mathbf w) + \lambda(T-t) \\
    &V(T, \mathbf w(T)) = \Phi(\mathbf w (T)) = 0
\end{align}
Finally, we consider the case without discounted rewards, i.e. $\gamma = 1$, which removes the $\log \gamma V$ term from \ref{eq:hjb} and drastically simplifies the equation:

\begin{align}
    &-\lambda  P(\mathbf w) - C(\mu^*) + \mu^* C'(\mu^*) = 0 \\
    \Rightarrow &\mu^* C'(\mu^*) - C(\mu^*) = \lambda - P(\mathbf w) \label{eq:cost_function_requirement}
\end{align}
The coefficient $\lambda$ can be determined from the boundary condition: the learning rate at the end of the episode should minimize the cost of learning, since any higher learning rate will only increase costs but not increase rewards. Thus, the learning rate at the end of the interval satisfies $\mu^*(T) = \arg\min C(\mu)$.
The equation above yields exact results for the optimal learning rate that, depending on the form of the cost $C(\mu)$ will either have a closed-form analytical solution or can be evaluated numerically. In this paper, we considered $C(\mu) =  \beta\mu^2$, thus the equations simplifies to $\beta {\mu^*}^2 = P(\mathbf w(T)) - P(\mathbf w (t))$, leading to the final expression:

\begin{equation}
    \mu^* = \sqrt{\frac{1}{\beta}(P(\mathbf w(T)) - P(\mathbf w(t))})
\end{equation}

\subsection{Estimating final performance through episodic memory}\label{sec:estimation_PT_supp}

We construct a performance estimator over the course of a learning episode for a new task by leveraging information from previous learning episodes. Each stored memory corresponds to a past learning trajectory, denoted by $\mathcal{T}_{h}(t)$, which is defined by performance over time:
\begin{equation}
\mathcal{T}_{h}(t) = \{P_{h}(0), P_{h}(dt), \ldots, P_{h}(t)\}.    
\end{equation}
When facing a new task, the agent compares its ongoing trajectory, $\mathcal{T}_{o}(t)$, with $H$ previously stored trajectories $\{\mathcal{T}_{h}(t)\}_{h \in H}$ retrieved from memory. Similarity-based weights $\{w_h(t)\}_{h \in H}$ are computed using a Gaussian kernel,
\begin{align}
k(\mathcal{T}_{o}(t), \mathcal{T}_{h}(t)) =
\exp\left(
-\frac{1}{2 N_{t} \sigma_{k}^{2}}
\sum_{i=0}^{N_{t}} \rho^{dt \cdot (N_{t} - i)}
\left[
P_{o}(dt \cdot i) - P_{h}(dt \cdot i)
\right]^{2}
\right), \label{eq:kernel_PT_est}
\end{align}
where $N_t$ is the number of discrete time steps stored in both trajectories up to time $t$, such that $N_t \cdot dt = t$, $\rho$ is a memory decay constant that assigns greater weight to the distances of more recent points along the trajectories, and $\sigma_{k}$ denotes the kernel width. The similarity weights are then normalized as
\begin{align}
w_h(t) =
\frac{
k(\mathcal{T}_{o}(t), \mathcal{T}_{h}(t))
}{
\sum_{h'=1}^{H}
k(\mathcal{T}_{o}(t), \mathcal{T}_{h'}(t))
}.
\end{align}
Before computing similarities, all trajectories are \textit{whitened} by subtracting their mean and dividing by their standard deviation, in order to avoid effects due to differences in performance scale. Using these similarity weights, future performance at times $t' > t$ can be extrapolated from memory as
\begin{equation}
\hat{P}(t' \mid t)
= \sum_{h =1}^{H} w_h(t) \, P_h(t'),
\quad \forall \, t < t' \leq T.
\end{equation}
To obtain the future performance estimate required for computing the optimal learning rate, this extrapolation is evaluated at $t' = T$, yielding $\hat{P}(T \mid t)$. Prior to observing any performance in a new task ($t = 0$), all similarity weights are set to $1/H$, resulting in the average trajectory over all stored memories.

\subsection{Computing the perceptron open-loop solution}\label{sec:perceptron_openloop}

We consider a linear perceptron with weights $\w(t)\in\mathbb{R}^d$ trained by gradient flow with a time-dependent learning rate $\mu(t)$,
\begin{equation}
    \frac{d\w}{dt} = -\mu(t)\nabla_{\w} L(\w),
\end{equation}
on a linear regression task with squared error loss
\begin{equation}
    L(\w) = \frac{1}{2}\left\langle (y-\w^\top \x)^2 \right\rangle,
\end{equation}
where inputs are drawn from an isotropic Gaussian $\x\sim\mathcal{N}(0,\sigma^2 I)$ and targets are generated by a teacher $\y=\w^{*\top}\x$. Under these assumptions, the gradient flow dynamics are exactly solvable and yield
\begin{equation}
    \w(t) = \w^* + (\w_0-\w^*) e^{-2\sigma^2 \int_0^t \mu(s)\,ds}.
\end{equation}
Defining the squared initialization distance $d^2=\|\w_0-\w^*\|^2$, the loss along the training trajectory becomes
\begin{equation}
    L(t) = \frac{\sigma^2}{2} d^2 \exp\!\left(-2\sigma^2 \int_0^t \mu(s)\,ds\right).
\end{equation}

The open-loop objective is to maximize the cumulative internal reward
\begin{equation}
    R(\mu) = \int_0^T \!\! dt \left[ L(t) + \beta \mu(t)^2 \right],
\end{equation}
where $\beta>0$ penalizes large learning rates. Taking the functional derivative of $R$ with respect to $\mu(\tau)$ yields an integro-differential optimality condition,
\begin{equation}
    \mu(\tau) = \frac{\sigma^2 d^2}{2\beta} \int_\tau^T \!\! dt \,
    \exp\!\left(-2\sigma^2 \int_0^t \mu(s)\,ds\right).
\end{equation}

To simplify this expression, we introduce the cumulative learning rate
\begin{equation}
    M(t) = \int_0^t \mu(s)\,ds,
\end{equation}
which transforms the optimality condition into a second-order ordinary differential equation,
\begin{equation}
    M''(t) + \frac{\sigma^2 d^2}{2\beta} e^{-2\sigma^2 M(t)} = 0.
\end{equation}
This equation admits a first integral,
\begin{equation}
    \mu(T)^2 - \mu(t)^2 = \frac{1}{\beta}\bigl(L(T)-L(t)\bigr),
\end{equation}
which coincides with the closed-loop solution derived earlier and highlights the consistency between the two formulations.

The equation can be solved in closed form by the substitution $u(t)=e^{-\sigma^2 M(t)}$, leading to an explicit expression for the optimal learning-rate schedule,
\begin{equation}
    \mu(t) = \theta \tan\!\bigl(\lambda\theta (T-t)\bigr),
\end{equation}
where $\theta$ is determined self-consistently by the terminal condition at $t=T$. The resulting transcendental equation for $\theta$ is solved numerically.

\subsection{Approximate solutions for the discount factor} \label{sec:homotopy_pade}

In this section, we solve the Hamilton-Jacobi-Bellman equation from Eq. \ref{eq:hjb} using a modified homotopy perturbation method \cite{saberi_nik_approximate-analytical_2012, ganjefar_modified_2016}. The nonlinear PDE we need to solve is obtained after substituting the internal reward rate, cost function and the gradient flow learning dynamics into Eq. \ref{eq:hjb} and maximizing the Hamiltonian:

\begin{align}
    &\log \gamma V(\mathbf w, t) + \frac{\partial V}{\partial t} + P(\mathbf w)  + \frac{1}{4\beta} \left( \frac{\partial V}{\partial \mathbf w}^\top  \frac{dP}{d\mathbf w} \right) ^2= 0 \\
    &V(\mathbf w(T), T) = 0 \\
    &\mu^*(t) = \frac{1}{2\beta} \frac{\partial V}{\partial \mathbf w}^\top  \frac{dP}{d\mathbf w} 
\end{align}

There is no known closed-loop solution for this PDE, but we find an analytical approximation. A homotopy is constructed with an embedding parameter $p\in[0,1]$ that continuously deforms an analytically tractable initial guess $V_g(\mathbf w,t)$, satisfying the terminal condition $V_g(\mathbf w,T)=0$, into the full nonlinear HJB equation \cite{saberi_nik_approximate-analytical_2012}.

The value function is assumed to admit a formal perturbation expansion
\begin{equation}
    V(\mathbf w,t;p) = \sum_{i=0}^{\infty} p^i V_i(\mathbf w,t),
\end{equation}
which, upon substitution into the homotopy and collection of terms of equal order in $p$, yields a hierarchy of linear first-order PDEs for the modes $V_i$. Each mode depends only on lower-order solutions and is solved subject to terminal conditions $V_i(\mathbf w,T)=0$, ensuring that any finite-order truncation satisfies the boundary condition. In practice, the series is truncated at low order and evaluated at $p=1$ to obtain an approximate solution to the original HJB equation.

To improve convergence of the perturbative solution at $p=1$, we apply Padé resummation to the truncated series, a modification from \cite{ganjefar_modified_2016}. Specifically, the power series in $p$ is replaced by a rational Padé approximant whose coefficients are chosen to match the Taylor expansion up to a given order. The final approximation to the value function and the optimal control $\mu^*(t)$ is obtained by evaluating the Padé approximant at $p=1$, which substantially enlarges the parameter regime over which the approximation remains well behaved.

Closed-form expressions for the resulting optimal control are derived for analytically tractable cases, including single-neuron linear regression and single-layer networks trained on a single task. The first mode of the homotopy-pade method applied to a single-neuron model is in Eq. \ref{eq:gamma_approximation}, with simulations in Fig. \ref{fig:open_loop}d,e.

\subsection{Implementation details} \label{sec:implementation_details}
All networks have been trained with an approximation of gradient flow via explicit Euler steps of size $dt$ in either full-batch or mini-batch gradient descent. The optimal learning rates from equation \ref{eq:closed_loop} are implemented using the online performance $P(\mathbf w(t))$, and the ground truth final performance $P(\mathbf w(T))$ obtained from a training episode with numerically-optimized learning rates (in all but the MNIST task). In all of the simulations presented here, performance is defined through the loss function, shifted so that performance is always positive $P(\mathbf w) = L(\mathbf w(0)) - L(\mathbf w)$. The shift was introduced purely for interpretability, it does not affect the results.

\subsubsection{Teacher-student}
The student network used in panels (a) and (b) of Figure \ref{fig:closed_loop} is a two-layer feedforward network with 4 inputs, 10 hidden units with biases, 6 output units without biases and $\tanh(\cdot)$ hidden layer activation function. The training set consists of 2048 samples with inputs were $\mathbf x_i \sim \mathcal N(\mathbf 0, I_4)$. Labels are produced by a teacher network with matching architecture, and weights drawn from a zero-mean Gaussian with standard variance scaling. The loss is the mean-squared error loss $L(\mathbf w) = \frac{1}{N}\sum_{i=1}^N \| \mathbf y_i - f(\mathbf x_i) \|^2$. The training algorithm is full-batch gradient descent, ran for 8000 steps with steps rescaled by $dt = 0.01$. Best performing baseline learning rate is $0.413$. The cost of learning was set to $\beta = 0.5$.

\subsubsection{MNIST}
The network used in panels (c) and (d) of Figure \ref{fig:closed_loop} is a two-layer feedforward network with 196 inputs, 100 hidden units with biases, 10 output units without biases, $\text{softplus}(\cdot)$ hidden layer activation function and a $\text{softmax}(\cdot)$ applied at the last layer. The network is trained on the image classification task MNIST \cite{deng_mnist_2012}. All of the input images have been downsampled by a factor of 2 prior to the network input. The loss is the cross-entropy loss $L(\mathbf w) = \sum_{i=1}^N \mathbf y_i^\top \log f(\mathbf x_i)$. Training is mini-batch gradient descent, ran for 3000 steps, with a 100 samples per batch per class and each update rescaled by $dt = 0.001$. Best baseline learning rate was found to be $0.172$. The ground truth final performance was found by finding the value of $P(T)$ that maximizes the cumulative reward when used in the closed-loop expression from Eq. \ref{eq:closed_loop}. It was found that $P(T) = 0.067$. The cost of learning was set to $\beta = 0.8$.

\subsubsection{Two-Gaussian discrimination} \label{sec:gaussian_discrimination}
In panels (e,f,g,h) in Fig. \ref{fig:closed_loop}, for simplicity all agents are  single neuron models performing a Two-Gaussian discrimination task via linear regression. Despite its simplicity, the learning rates and weight dynamics remain similar to the complex MNIST classification task. The task outputs are either $y_i=1$ or $y_i=-1$ with equal probability, and inputs are sampled from the appropriate Gaussian $x_i \sim \mathcal{N}(yd, \sigma^2)$, where $d = 2$, $\sigma^2 = 4$. Training is a full-batch gradient descent with rescaled updates by $dt = 0.001$, ran for 700 steps. The cost of learning was set to $\beta = 0.5$.

\subsubsection{Figure 3}
A nonlinear two-layer (student) network with a softplus activation function and six hidden neurons is trained to map samples from a Gaussian input,
$\mathbf{x}_i \sim \mathcal{N}(\mathbf{0}, I_3)$, to targets $y \in \mathbb{R}^2$ generated by a nonlinear two-layer (teacher) network.
Training is performed by minimizing the mean squared error using backpropagation, with stochastic gradient descent updates, a base learning rate $\alpha=10.0$, and random batches of size 32 for each update. Each of the 1000 learning instances shown in the figure corresponds to the training of a completely new student--teacher pair.
Teacher weights are sampled using the standard Xavier initialization scheme, while student network weights are sampled from a Gaussian distribution
with standard deviation $0.001$, in order to induce rich nonlinear learning dynamics in the student across learning instances.
To initialize the meta-training across many learning instances, the base learning rate is used to generate an initial learning trajectory.
For all other learning instances, we use a modified version of the closed-loop expression in Eq.~\ref{eq:closed_loop}, in which the base learning rate update is given by
\begin{align}
    \frac{d\w}{dt} = -(\mu(t) + \alpha)\nabla_{\w} L(\w),
\end{align}
and the optimal learning rate schedule is
\begin{align}
    \mu^* = - \alpha + \sqrt{\frac{1}{\beta}\bigl(P(\mathbf w(T)) - P(\mathbf w(t))\bigr) + \alpha^2},
\end{align}
with a cost function $C(\mu) = \beta \mu^2$. This formulation is used to model passive learning that does not incur a cost, and qualitatively similar results are obtained for $\alpha = 0$.
To apply the closed-form solution, we estimate the value of $P(T)$ for each learning instance using the episodic memory method described in
\autoref{sec:estimating_final_performance} and Methods~\ref{sec:estimation_PT_supp}. Each training instance consists of $N_t = 5000$ gradient steps, with memory decay $\rho = 0.7$, kernel width $\sigma_k = 0.1$ in
Eq.~\ref{eq:kernel_PT_est}, $dt = 0.001$, $\beta=0.01$ and $\gamma = 1.0$.
Panel (a) shows the predicted future performance after 500 steps of the inner loop (vertical dashed line) using 25 learning instances in memory.
Panel (b) shows the same analysis but predicting future performance after 3000 inner-loop steps.
Panel (c) shows the 25th and 75th percentiles of $\hat{P}(T \mid t) - P(T)$ for different numbers of learning instances in memory.
Panel (d) shows $\hat{P}(T \mid t) - P(T)$ versus the standard deviation of $P(t)$ across time for each of the 1000 learning instances.

\subsubsection{Figure 4}
Panel (a): linear perceptron of input dimension 4 is trained on a teacher-student regression, with a randomly initialized and matched teacher architecture. Training is performed with a 1000 steps rescaled by $dt=0.01$, with $\beta = 0.1$, $\gamma=1$. Panels (b,c) are simulated analytical expressions where in (b) cost of effort $\beta$ is varied between 0.2 and 3.0, in (c) task difficulty $d$ is varied between 0.1 and 1.0. Panels (d, e) show a linear neuron learning a linear regression task, training with 500 steps rescaled by dt=0.001. Panel (d) shows numerically optimized learning rates, panel (e) shows learning rates obtained the expression from Eq. \ref{eq:gamma_approximation}. Discount factor $\gamma$ varied between 0.1 and 0.95. Cost of effort is $\beta = 5$.  Panel (f): student-teacher task, student is a two-layer network with 4 inputs, 4 hidden neurons, 6 outputs, initialized with small weights; teacher is a randomly initialized two-layer network with 4 inputs, 10 hidden neurons, 6 outputs, both with $\text{tanh}(\cdot)$ nonlinearity. Training is run for 8000 steps with $dt = 0.01$, cost of effort is $\beta=0.1$. Discount factor $\gamma$ is varied between 0.1 and 1.0.

\subsubsection{Figure 5}
Left panel is adapted from Masis et al \cite{masis_strategically_2023}, Fig. 7b. Right panel shows a single linear neuron trained for a 130 steps with $dt=0.001$ on a linear regression task. Cost of effort was $\beta=0.05$. Optimal learning rate obtained from Eq. \ref{eq:closed_loop}, the suboptimal (low) learning speed has the optimal shape but scaled down by 0.6 compared to the optimal one.

\subsection{Numerical optimization}
\label{sec:numerical_methods}

All numerically optimized learning-rate schedules are obtained using a meta-optimization procedure adapted from \cite{carrasco-davis_meta-learning_2024}. The learning rate function $\mu(t)$ is optimized directly by maximizing the cumulative internal reward $R(\mu)$. Gradients of $R$ with respect to $\mu$ are computed by differentiating through the full training dynamics using automatic differentiation (implemented in \texttt{JAX}). At each meta-iteration, $\mu$ is updated by gradient ascent in function space until convergence to a locally optimal solution. This procedure yields numerically optimized open-loop learning-rate schedules without relying on analytical approximations.

\section*{Acknowledgments} 
We thank Javier Masís for useful discussions. We thank the following funding sources: Gatsby Charitable Foundation (GAT3850 and GAT4058) to VN, RC, AS, and PEL; Sainsbury Wellcome Centre Core Grant from Wellcome (219627/Z/19/Z) to
AS; Schmidt Science Polymath Award to AS. AS is a CIFAR Azrieli Global Scholar in the Learning
in Machines \& Brains program. RC thanks the Princeton Presidential Fellowship.

\newpage

\bibliographystyle{unsrtnat}
\bibliography{references} 

\begin{thebibliography}{118}
\providecommand{\natexlab}[1]{#1}
\providecommand{\url}[1]{\texttt{#1}}
\expandafter\ifx\csname urlstyle\endcsname\relax
  \providecommand{\doi}[1]{doi: #1}\else
  \providecommand{\doi}{doi: \begingroup \urlstyle{rm}\Url}\fi

\bibitem[Taffoni et~al.(2014)Taffoni, Tamilia, Focaroli, Formica, Ricci, Di~Pino, Baldassarre, Mirolli, Guglielmelli, and Keller]{taffoni_development_2014}
Fabrizio Taffoni, Eleonora Tamilia, Valentina Focaroli, Domenico Formica, Luca Ricci, Giovanni Di~Pino, Gianluca Baldassarre, Marco Mirolli, Eugenio Guglielmelli, and Flavio Keller.
\newblock Development of goal-directed action selection guided by intrinsic motivations: an experiment with children.
\newblock \emph{Experimental Brain Research}, 232\penalty0 (7):\penalty0 2167--2177, July 2014.
\newblock ISSN 1432-1106.
\newblock \doi{10.1007/s00221-014-3907-z}.
\newblock URL \url{https://doi.org/10.1007/s00221-014-3907-z}.

\bibitem[Raz and Saxe(2020)]{raz_learning_2020}
Gal Raz and Rebecca Saxe.
\newblock Learning in {Infancy} {Is} {Active}, {Endogenously} {Motivated}, and {Depends} on the {Prefrontal} {Cortices}.
\newblock \emph{Annual Review of Developmental Psychology}, 2\penalty0 (Volume 2, 2020):\penalty0 247--268, December 2020.
\newblock ISSN 2640-7922.
\newblock \doi{10.1146/annurev-devpsych-121318-084841}.
\newblock URL \url{https://www.annualreviews.org/content/journals/10.1146/annurev-devpsych-121318-084841}.
\newblock Publisher: Annual Reviews.

\bibitem[van Gog et~al.(2020)van Gog, Hoogerheide, and van Harsel]{van_gog_role_2020}
Tamara van Gog, Vincent Hoogerheide, and Milou van Harsel.
\newblock The {Role} of {Mental} {Effort} in {Fostering} {Self}-{Regulated} {Learning} with {Problem}-{Solving} {Tasks}.
\newblock \emph{Educational Psychology Review}, 32\penalty0 (4):\penalty0 1055--1072, December 2020.
\newblock ISSN 1573-336X.
\newblock \doi{10.1007/s10648-020-09544-y}.
\newblock URL \url{https://doi.org/10.1007/s10648-020-09544-y}.

\bibitem[Kidd et~al.(2012)Kidd, Piantadosi, and Aslin]{kidd_goldilocks_2012}
Celeste Kidd, Steven~T. Piantadosi, and Richard~N. Aslin.
\newblock The {Goldilocks} {Effect}: {Human} {Infants} {Allocate} {Attention} to {Visual} {Sequences} {That} {Are} {Neither} {Too} {Simple} {Nor} {Too} {Complex}.
\newblock \emph{PLOS ONE}, 7\penalty0 (5):\penalty0 e36399, May 2012.
\newblock ISSN 1932-6203.
\newblock \doi{10.1371/journal.pone.0036399}.
\newblock URL \url{https://journals.plos.org/plosone/article?id=10.1371/journal.pone.0036399}.
\newblock Publisher: Public Library of Science.

\bibitem[Bonawitz et~al.(2018)Bonawitz, Bass, and Lapidow]{bonawitz_choosing_2018}
Elizabeth Bonawitz, Ilona Bass, and Elizabeth Lapidow.
\newblock Choosing to {Learn}: {Evidence} {Evaluation} for {Active} {Learning} and {Teaching} in {Early} {Childhood}.
\newblock In Megan~M. Saylor and Patricia~A. Ganea, editors, \emph{Active {Learning} from {Infancy} to {Childhood}: {Social} {Motivation}, {Cognition}, and {Linguistic} {Mechanisms}}, pages 213--231. Springer International Publishing, Cham, 2018.
\newblock ISBN 978-3-319-77182-3.
\newblock \doi{10.1007/978-3-319-77182-3_12}.
\newblock URL \url{https://doi.org/10.1007/978-3-319-77182-3_12}.

\bibitem[Castro et~al.(2008)Castro, Kalish, Nowak, Qian, Rogers, and Zhu]{castro_human_2008}
Rui Castro, Charles Kalish, Robert Nowak, Ruichen Qian, Tim Rogers, and Jerry Zhu.
\newblock Human {Active} {Learning}.
\newblock In \emph{Advances in {Neural} {Information} {Processing} {Systems}}, volume~21. Curran Associates, Inc., 2008.
\newblock URL \url{https://papers.nips.cc/paper_files/paper/2008/hash/fc49306d97602c8ed1be1dfbf0835ead-Abstract.html}.

\bibitem[Markant and Gureckis(2014)]{markant_is_2014}
Douglas~B. Markant and Todd~M. Gureckis.
\newblock Is it better to select or to receive? {Learning} via active and passive hypothesis testing.
\newblock \emph{Journal of Experimental Psychology: General}, 143\penalty0 (1):\penalty0 94--122, 2014.
\newblock ISSN 1939-2222.
\newblock \doi{10.1037/a0032108}.
\newblock Place: US Publisher: American Psychological Association.

\bibitem[Coenen et~al.(2019)Coenen, Nelson, and Gureckis]{coenen_asking_2019}
Anna Coenen, Jonathan~D. Nelson, and Todd~M. Gureckis.
\newblock Asking the right questions about the psychology of human inquiry: {Nine} open challenges.
\newblock \emph{Psychonomic Bulletin \& Review}, 26\penalty0 (5):\penalty0 1548--1587, October 2019.
\newblock ISSN 1531-5320.
\newblock \doi{10.3758/s13423-018-1470-5}.
\newblock URL \url{https://doi.org/10.3758/s13423-018-1470-5}.

\bibitem[Tullis and Benjamin(2011)]{tullis_effectiveness_2011}
Jonathan~G. Tullis and Aaron~S. Benjamin.
\newblock On the effectiveness of self-paced learning.
\newblock \emph{Journal of Memory and Language}, 64\penalty0 (2):\penalty0 109--118, February 2011.
\newblock ISSN 0749-596X.
\newblock \doi{10.1016/j.jml.2010.11.002}.
\newblock URL \url{https://www.sciencedirect.com/science/article/pii/S0749596X10000999}.

\bibitem[Son(2010)]{son_metacognitive_2010}
Lisa~K. Son.
\newblock Metacognitive control and the spacing effect.
\newblock \emph{Journal of Experimental Psychology: Learning, Memory, and Cognition}, 36\penalty0 (1):\penalty0 255--262, 2010.
\newblock ISSN 1939-1285, 0278-7393.
\newblock \doi{10.1037/a0017892}.
\newblock URL \url{https://doi.apa.org/doi/10.1037/a0017892}.

\bibitem[Bazhydai et~al.(2020)Bazhydai, Twomey, and Westermann]{bazhydai_curiosity_2020}
Marina Bazhydai, Katherine Twomey, and Gert Westermann.
\newblock Curiosity and {Exploration}.
\newblock In Janette~B. Benson, editor, \emph{Encyclopedia of {Infant} and {Early} {Childhood} {Development} ({Second} {Edition})}, pages 370--378. Elsevier, Oxford, January 2020.
\newblock ISBN 978-0-12-816511-9.
\newblock \doi{10.1016/B978-0-12-809324-5.05804-1}.
\newblock URL \url{https://www.sciencedirect.com/science/article/pii/B9780128093245058041}.

\bibitem[Hidi and Renninger(2019)]{hidi_interest_2019}
Suzanne~E. Hidi and K.~Ann Renninger.
\newblock Interest {Development} and {Its} {Relation} to {Curiosity}: {Needed} {Neuroscientific} {Research}.
\newblock \emph{Educational Psychology Review}, 31\penalty0 (4):\penalty0 833--852, December 2019.
\newblock ISSN 1573-336X.
\newblock \doi{10.1007/s10648-019-09491-3}.
\newblock URL \url{https://doi.org/10.1007/s10648-019-09491-3}.

\bibitem[Schwartenbeck et~al.(2019)Schwartenbeck, Passecker, Hauser, FitzGerald, Kronbichler, and Friston]{schwartenbeck_computational_2019}
Philipp Schwartenbeck, Johannes Passecker, Tobias~U Hauser, Thomas~HB FitzGerald, Martin Kronbichler, and Karl~J Friston.
\newblock Computational mechanisms of curiosity and goal-directed exploration.
\newblock \emph{eLife}, 8:\penalty0 e41703, May 2019.
\newblock ISSN 2050-084X.
\newblock \doi{10.7554/eLife.41703}.
\newblock URL \url{https://doi.org/10.7554/eLife.41703}.
\newblock Publisher: eLife Sciences Publications, Ltd.

\bibitem[Geana et~al.(2016)Geana, Wilson, Daw, and Cohen]{geana_boredom_2016}
Andra Geana, Robert~C. Wilson, Nathaniel Daw, and Jonathan~D. Cohen.
\newblock Boredom, {Information}-{Seeking} and {Exploration}.
\newblock In \emph{Proceedings of the 38th {Annual} {Meeting} of the {Cognitive} {Science} {Society}, {CogSci} 2016}, pages 1751--1756. The Cognitive Science Society, 2016.
\newblock URL \url{https://collaborate.princeton.edu/en/publications/boredom-information-seeking-and-exploration}.

\bibitem[Agrawal et~al.(2022)Agrawal, Mattar, Cohen, and Daw]{agrawal_temporal_2022}
Mayank Agrawal, Marcelo~G. Mattar, Jonathan~D. Cohen, and Nathaniel~D. Daw.
\newblock The temporal dynamics of opportunity costs: {A} normative account of cognitive fatigue and boredom.
\newblock \emph{Psychological Review}, 129\penalty0 (3):\penalty0 564--585, April 2022.
\newblock ISSN 1939-1471.
\newblock \doi{10.1037/rev0000309}.

\bibitem[Baranes et~al.(2014)Baranes, Oudeyer, and Gottlieb]{baranes_effects_2014}
Adrien~F. Baranes, Pierre-Yves Oudeyer, and Jacqueline Gottlieb.
\newblock The effects of task difficulty, novelty and the size of the search space on intrinsically motivated exploration.
\newblock \emph{Frontiers in Neuroscience}, 8, October 2014.
\newblock ISSN 1662-453X.
\newblock \doi{10.3389/fnins.2014.00317}.
\newblock URL \url{https://www.frontiersin.org/journals/neuroscience/articles/10.3389/fnins.2014.00317/full}.
\newblock Publisher: Frontiers.

\bibitem[Schulz et~al.(2019)Schulz, Bhui, Love, Brier, Todd, and Gershman]{schulz_structured_2019}
Eric Schulz, Rahul Bhui, Bradley~C. Love, Bastien Brier, Michael~T. Todd, and Samuel~J. Gershman.
\newblock Structured, uncertainty-driven exploration in real-world consumer choice.
\newblock \emph{Proceedings of the National Academy of Sciences}, 116\penalty0 (28):\penalty0 13903--13908, July 2019.
\newblock \doi{10.1073/pnas.1821028116}.
\newblock URL \url{https://www.pnas.org/doi/10.1073/pnas.1821028116}.
\newblock Publisher: Proceedings of the National Academy of Sciences.

\bibitem[Wilson et~al.(2019)Wilson, Shenhav, Straccia, and Cohen]{wilson_eighty_2019}
Robert~C. Wilson, Amitai Shenhav, Mark Straccia, and Jonathan~D. Cohen.
\newblock The {Eighty} {Five} {Percent} {Rule} for optimal learning.
\newblock \emph{Nature Communications}, 10\penalty0 (1):\penalty0 4646, November 2019.
\newblock ISSN 2041-1723.
\newblock \doi{10.1038/s41467-019-12552-4}.
\newblock URL \url{https://www.nature.com/articles/s41467-019-12552-4}.
\newblock Publisher: Nature Publishing Group.

\bibitem[Ten et~al.(2021)Ten, Kaushik, Oudeyer, and Gottlieb]{ten_humans_2021}
Alexandr Ten, Pramod Kaushik, Pierre-Yves Oudeyer, and Jacqueline Gottlieb.
\newblock Humans monitor learning progress in curiosity-driven exploration.
\newblock \emph{Nature Communications}, 12\penalty0 (1):\penalty0 5972, October 2021.
\newblock ISSN 2041-1723.
\newblock \doi{10.1038/s41467-021-26196-w}.
\newblock URL \url{https://www.nature.com/articles/s41467-021-26196-w}.
\newblock Publisher: Nature Publishing Group.

\bibitem[Cubit et~al.(2021)Cubit, Canale, Handsman, Kidd, and Bennetto]{cubit_visual_2021}
Laura~S. Cubit, Rebecca Canale, Rebecca Handsman, Celeste Kidd, and Loisa Bennetto.
\newblock Visual {Attention} {Preference} for {Intermediate} {Predictability} in {Young} {Children}.
\newblock \emph{Child Development}, 92\penalty0 (2):\penalty0 691--703, March 2021.
\newblock ISSN 1467-8624.
\newblock \doi{10.1111/cdev.13536}.

\bibitem[Son and Metcalfe(2000)]{son_metacognitive_2000}
Lisa~K. Son and Janet Metcalfe.
\newblock Metacognitive and control strategies in study-time allocation.
\newblock \emph{Journal of Experimental Psychology: Learning, Memory, and Cognition}, 26\penalty0 (1):\penalty0 204--221, 2000.
\newblock ISSN 1939-1285.
\newblock \doi{10.1037/0278-7393.26.1.204}.
\newblock Place: US Publisher: American Psychological Association.

\bibitem[de~Eccher et~al.(2024)de~Eccher, Mundry, and Mani]{de_eccher_childrens_2024}
Martina de~Eccher, Roger Mundry, and Nivedita Mani.
\newblock Children’s subjective uncertainty-driven sampling behaviour.
\newblock \emph{Royal Society Open Science}, 11\penalty0 (4):\penalty0 231283, April 2024.
\newblock \doi{10.1098/rsos.231283}.
\newblock URL \url{https://royalsocietypublishing.org/doi/full/10.1098/rsos.231283}.
\newblock Publisher: Royal Society.

\bibitem[Masís et~al.(2023)Masís, Chapman, Rhee, Cox, and Saxe]{masis_strategically_2023}
Javier Masís, Travis Chapman, Juliana~Y Rhee, David~D Cox, and Andrew~M Saxe.
\newblock Strategically managing learning during perceptual decision making.
\newblock \emph{eLife}, 12:\penalty0 e64978, February 2023.
\newblock ISSN 2050-084X.
\newblock \doi{10.7554/eLife.64978}.
\newblock URL \url{https://doi.org/10.7554/eLife.64978}.
\newblock Publisher: eLife Sciences Publications, Ltd.

\bibitem[Obando et~al.(2025)Obando, Musslick, and Cohen]{obando_learning_2025}
Javier Alejandro~Masís Obando, Sebastian Musslick, and Jonathan~D. Cohen.
\newblock Learning expectations shape cognitive control allocation.
\newblock \emph{Proceedings of the National Academy of Sciences}, 122\penalty0 (44):\penalty0 e2416720122, November 2025.
\newblock \doi{10.1073/pnas.2416720122}.
\newblock URL \url{https://www.pnas.org/doi/10.1073/pnas.2416720122}.
\newblock Publisher: Proceedings of the National Academy of Sciences.

\bibitem[Potter et~al.(2010)Potter, O'Riordan, Barnett, Osting, Wagoner, Burger, and Roopra]{potter_metabolic_2010}
Wyatt~B. Potter, Kenneth~J. O'Riordan, David Barnett, Susan M.~K. Osting, Matthew Wagoner, Corinna Burger, and Avtar Roopra.
\newblock Metabolic regulation of neuronal plasticity by the energy sensor {AMPK}.
\newblock \emph{PloS One}, 5\penalty0 (2):\penalty0 e8996, February 2010.
\newblock ISSN 1932-6203.
\newblock \doi{10.1371/journal.pone.0008996}.

\bibitem[Smith et~al.(2011)Smith, Riby, Eekelen, and Foster]{smith_glucose_2011}
Michael~A. Smith, Leigh~M. Riby, J.~Anke M.~van Eekelen, and Jonathan~K. Foster.
\newblock Glucose enhancement of human memory: a comprehensive research review of the glucose memory facilitation effect.
\newblock \emph{Neuroscience and Biobehavioral Reviews}, 35\penalty0 (3):\penalty0 770--783, January 2011.
\newblock ISSN 1873-7528.
\newblock \doi{10.1016/j.neubiorev.2010.09.008}.

\bibitem[Klug et~al.(2022)Klug, Godbersen, Rischka, Wadsak, Pichler, Klöbl, Hacker, Lanzenberger, and Hahn]{klug_learning_2022}
Sebastian Klug, Godber~M. Godbersen, Lucas Rischka, Wolfgang Wadsak, Verena Pichler, Manfred Klöbl, Marcus Hacker, Rupert Lanzenberger, and Andreas Hahn.
\newblock Learning induces coordinated neuronal plasticity of metabolic demands and functional brain networks.
\newblock \emph{Communications Biology}, 5\penalty0 (1):\penalty0 428, May 2022.
\newblock ISSN 2399-3642.
\newblock \doi{10.1038/s42003-022-03362-4}.
\newblock URL \url{https://www.nature.com/articles/s42003-022-03362-4}.
\newblock Publisher: Nature Publishing Group.

\bibitem[Mery and Kawecki(2004)]{mery_operating_2004}
Frederic Mery and Tadeusz~J. Kawecki.
\newblock An operating cost of learning in \textit{{Drosophila} melanogaster}.
\newblock \emph{Animal Behaviour}, 68\penalty0 (3):\penalty0 589--598, September 2004.
\newblock ISSN 0003-3472.
\newblock \doi{10.1016/j.anbehav.2003.12.005}.
\newblock URL \url{https://www.sciencedirect.com/science/article/pii/S0003347204001708}.

\bibitem[Snell-Rood et~al.(2011)Snell-Rood, Davidowitz, and Papaj]{snell-rood_reproductive_2011}
Emilie~C. Snell-Rood, Goggy Davidowitz, and Daniel~R. Papaj.
\newblock Reproductive tradeoffs of learning in a butterfly.
\newblock \emph{Behavioral Ecology}, 22\penalty0 (2):\penalty0 291--302, March 2011.
\newblock ISSN 1045-2249.
\newblock \doi{10.1093/beheco/arq169}.
\newblock URL \url{https://doi.org/10.1093/beheco/arq169}.

\bibitem[Mery and Kawecki(2005)]{mery_cost_2005}
Frederic Mery and Tadeusz~J. Kawecki.
\newblock A {Cost} of {Long}-{Term} {Memory} in {Drosophila}.
\newblock \emph{Science}, 308\penalty0 (5725):\penalty0 1148--1148, May 2005.
\newblock \doi{10.1126/science.1111331}.
\newblock URL \url{https://www.science.org/doi/10.1126/science.1111331}.
\newblock Publisher: American Association for the Advancement of Science.

\bibitem[Masís et~al.(2021)Masís, Musslick, and Cohen]{masis_value_2021}
Javier~Alejandro Masís, Sebastian Musslick, and Jonathan Cohen.
\newblock The {Value} of {Learning} and {Cognitive} {Control} {Allocation}.
\newblock \emph{Proceedings of the Annual Meeting of the Cognitive Science Society}, 2021.
\newblock URL \url{https://escholarship.org/uc/item/7w0223v0}.

\bibitem[Masis et~al.(2024)Masis, Musslick, and Cohen]{masis_learning_2024}
Javier~Alejandro Masis, Sebastian Musslick, and Jonathan~D. Cohen.
\newblock Learning expectations shape cognitive control allocation, August 2024.
\newblock URL \url{https://osf.io/d2cbg_v1}.

\bibitem[Carrasco-Davis et~al.(2024)Carrasco-Davis, Masís, and Saxe]{carrasco-davis_meta-learning_2024}
Rodrigo Carrasco-Davis, Javier Masís, and Andrew~M. Saxe.
\newblock Meta-{Learning} {Strategies} through {Value} {Maximization} in {Neural} {Networks}, July 2024.
\newblock URL \url{http://arxiv.org/abs/2310.19919}.
\newblock arXiv:2310.19919 [cs].

\bibitem[Kool and Botvinick(2018)]{kool_mental_2018}
Wouter Kool and Matthew Botvinick.
\newblock Mental labour.
\newblock \emph{Nature Human Behaviour}, 2\penalty0 (12):\penalty0 899--908, December 2018.
\newblock ISSN 2397-3374.
\newblock \doi{10.1038/s41562-018-0401-9}.
\newblock URL \url{https://www.nature.com/articles/s41562-018-0401-9}.
\newblock Publisher: Nature Publishing Group.

\bibitem[Kurzban et~al.(2013)Kurzban, Duckworth, Kable, and Myers]{kurzban_opportunity_2013}
Robert Kurzban, Angela Duckworth, Joseph~W. Kable, and Justus Myers.
\newblock An opportunity cost model of subjective effort and task performance.
\newblock \emph{The Behavioral and brain sciences}, 36\penalty0 (6):\penalty0 10.1017/S0140525X12003196, December 2013.
\newblock ISSN 0140-525X.
\newblock \doi{10.1017/S0140525X12003196}.
\newblock URL \url{https://www.ncbi.nlm.nih.gov/pmc/articles/PMC3856320/}.

\bibitem[Shenhav et~al.(2013)Shenhav, Botvinick, and Cohen]{shenhav_expected_2013}
Amitai Shenhav, Matthew~M. Botvinick, and Jonathan~D. Cohen.
\newblock The {Expected} {Value} of {Control}: {An} {Integrative} {Theory} of {Anterior} {Cingulate} {Cortex} {Function}.
\newblock \emph{Neuron}, 79\penalty0 (2):\penalty0 217--240, July 2013.
\newblock ISSN 0896-6273.
\newblock \doi{10.1016/j.neuron.2013.07.007}.
\newblock URL \url{https://www.sciencedirect.com/science/article/pii/S0896627313006077}.

\bibitem[Shenhav et~al.(2017)Shenhav, Musslick, Lieder, Kool, Griffiths, Cohen, and Botvinick]{shenhav_toward_2017}
Amitai Shenhav, Sebastian Musslick, Falk Lieder, Wouter Kool, Thomas~L. Griffiths, Jonathan~D. Cohen, and Matthew~M. Botvinick.
\newblock Toward a {Rational} and {Mechanistic} {Account} of {Mental} {Effort}.
\newblock \emph{Annual Review of Neuroscience}, 40:\penalty0 99--124, July 2017.
\newblock ISSN 1545-4126.
\newblock \doi{10.1146/annurev-neuro-072116-031526}.

\bibitem[Jarvis et~al.(2022)Jarvis, Stevenson, Huynh, Babbage, Coxon, and Chong]{jarvis_effort_2022}
Huw Jarvis, Isabelle Stevenson, Amy~Q. Huynh, Emily Babbage, James Coxon, and Trevor T.-J. Chong.
\newblock Effort {Reinforces} {Learning}.
\newblock \emph{Journal of Neuroscience}, 42\penalty0 (40):\penalty0 7648--7658, October 2022.
\newblock ISSN 0270-6474, 1529-2401.
\newblock \doi{10.1523/JNEUROSCI.2223-21.2022}.
\newblock URL \url{https://www.jneurosci.org/content/42/40/7648}.
\newblock Publisher: Society for Neuroscience Section: Research Articles.

\bibitem[Sayalı et~al.(2023)Sayalı, Heling, and Cools]{sayali_learning_2023}
Ceyda Sayalı, Emma Heling, and Roshan Cools.
\newblock Learning progress mediates the link between cognitive effort and task engagement.
\newblock \emph{Cognition}, 236:\penalty0 105418, July 2023.
\newblock ISSN 0010-0277.
\newblock \doi{10.1016/j.cognition.2023.105418}.
\newblock URL \url{https://www.sciencedirect.com/science/article/pii/S0010027723000525}.

\bibitem[Hospedales et~al.(2020)Hospedales, Antoniou, Micaelli, and Storkey]{hospedales_meta-learning_2020}
Timothy Hospedales, Antreas Antoniou, Paul Micaelli, and Amos Storkey.
\newblock Meta-{Learning} in {Neural} {Networks}: {A} {Survey}, November 2020.
\newblock URL \url{http://arxiv.org/abs/2004.05439}.
\newblock arXiv:2004.05439 [cs, stat].

\bibitem[Zhang et~al.(2021)Zhang, Song, Yao, and Gao]{zhang_curriculum-based_2021}
Ji~Zhang, Jingkuan Song, Yazhou Yao, and Lianli Gao.
\newblock Curriculum-{Based} {Meta}-learning.
\newblock In \emph{Proceedings of the 29th {ACM} {International} {Conference} on {Multimedia}}, {MM} '21, pages 1838--1846, New York, NY, USA, October 2021. Association for Computing Machinery.
\newblock ISBN 978-1-4503-8651-7.
\newblock \doi{10.1145/3474085.3475335}.
\newblock URL \url{https://doi.org/10.1145/3474085.3475335}.

\bibitem[Stergiadis et~al.(2021)Stergiadis, Agrawal, and Squire]{stergiadis_curriculum_2021}
Emmanouil Stergiadis, Priyanka Agrawal, and Oliver Squire.
\newblock Curriculum {Meta}-{Learning} for {Few}-shot {Classification}, December 2021.
\newblock URL \url{http://arxiv.org/abs/2112.02913}.
\newblock arXiv:2112.02913 [cs].

\bibitem[Soviany et~al.(2022)Soviany, Ionescu, Rota, and Sebe]{soviany_curriculum_2022}
Petru Soviany, Radu~Tudor Ionescu, Paolo Rota, and Nicu Sebe.
\newblock Curriculum {Learning}: {A} {Survey}, April 2022.
\newblock URL \url{http://arxiv.org/abs/2101.10382}.
\newblock arXiv:2101.10382 [cs].

\bibitem[Finn et~al.(2017)Finn, Abbeel, and Levine]{finn_model-agnostic_2017}
Chelsea Finn, Pieter Abbeel, and Sergey Levine.
\newblock Model-{Agnostic} {Meta}-{Learning} for {Fast} {Adaptation} of {Deep} {Networks}, July 2017.
\newblock URL \url{http://arxiv.org/abs/1703.03400}.
\newblock arXiv:1703.03400 [cs].

\bibitem[Parisi et~al.(2019)Parisi, Kemker, Part, Kanan, and Wermter]{parisi_continual_2019}
German~I. Parisi, Ronald Kemker, Jose~L. Part, Christopher Kanan, and Stefan Wermter.
\newblock Continual {Lifelong} {Learning} with {Neural} {Networks}: {A} {Review}.
\newblock \emph{Neural Networks}, 113:\penalty0 54--71, May 2019.
\newblock ISSN 08936080.
\newblock \doi{10.1016/j.neunet.2019.01.012.}
\newblock URL \url{http://arxiv.org/abs/1802.07569}.
\newblock arXiv:1802.07569 [cs].

\bibitem[Wang et~al.(2024)Wang, Zhang, Su, and Zhu]{wang_comprehensive_2024}
Liyuan Wang, Xingxing Zhang, Hang Su, and Jun Zhu.
\newblock A {Comprehensive} {Survey} of {Continual} {Learning}: {Theory}, {Method} and {Application}.
\newblock \emph{IEEE Transactions on Pattern Analysis and Machine Intelligence}, 46\penalty0 (8):\penalty0 5362--5383, August 2024.
\newblock ISSN 1939-3539.
\newblock \doi{10.1109/TPAMI.2024.3367329}.
\newblock URL \url{https://ieeexplore.ieee.org/document/10444954/?arnumber=10444954}.
\newblock Conference Name: IEEE Transactions on Pattern Analysis and Machine Intelligence.

\bibitem[Franceschi et~al.(2018)Franceschi, Frasconi, Salzo, Grazzi, and Pontil]{franceschi_bilevel_2018}
Luca Franceschi, Paolo Frasconi, Saverio Salzo, Riccardo Grazzi, and Massimilano Pontil.
\newblock Bilevel {Programming} for {Hyperparameter} {Optimization} and {Meta}-{Learning}, July 2018.
\newblock URL \url{http://arxiv.org/abs/1806.04910}.
\newblock arXiv:1806.04910 [cs, stat].

\bibitem[Baik et~al.(2020)Baik, Choi, Choi, Kim, and Lee]{baik_meta-learning_2020}
Sungyong Baik, Myungsub Choi, Janghoon Choi, Heewon Kim, and Kyoung~Mu Lee.
\newblock Meta-{Learning} with {Adaptive} {Hyperparameters}.
\newblock In \emph{Advances in {Neural} {Information} {Processing} {Systems}}, volume~33, pages 20755--20765. Curran Associates, Inc., 2020.
\newblock URL \url{https://proceedings.neurips.cc//paper/2020/hash/ee89223a2b625b5152132ed77abbcc79-Abstract.html}.

\bibitem[Nakamura et~al.(2021)Nakamura, Derbel, Won, and Hong]{nakamura_learning-rate_2021}
Kensuke Nakamura, Bilel Derbel, Kyoung-Jae Won, and Byung-Woo Hong.
\newblock Learning-{Rate} {Annealing} {Methods} for {Deep} {Neural} {Networks}.
\newblock \emph{Electronics}, 10\penalty0 (16):\penalty0 2029, January 2021.
\newblock ISSN 2079-9292.
\newblock \doi{10.3390/electronics10162029}.
\newblock URL \url{https://www.mdpi.com/2079-9292/10/16/2029}.
\newblock Number: 16 Publisher: Multidisciplinary Digital Publishing Institute.

\bibitem[Mori et~al.(2025)Mori, Mannelli, and Mignacco]{mori_optimal_2025}
Francesco Mori, Stefano~Sarao Mannelli, and Francesca Mignacco.
\newblock Optimal {Protocols} for {Continual} {Learning} via {Statistical} {Physics} and {Control} {Theory}, May 2025.
\newblock URL \url{http://arxiv.org/abs/2409.18061}.
\newblock arXiv:2409.18061 [cs].

\bibitem[Mignacco and Mori(2025)]{mignacco_statistical_2025}
Francesca Mignacco and Francesco Mori.
\newblock A statistical physics framework for optimal learning, July 2025.
\newblock URL \url{http://arxiv.org/abs/2507.07907}.
\newblock arXiv:2507.07907 [cond-mat].

\bibitem[Saxe et~al.(2019)Saxe, McClelland, and Ganguli]{saxe_mathematical_2019}
Andrew~M. Saxe, James~L. McClelland, and Surya Ganguli.
\newblock A mathematical theory of semantic development in deep neural networks.
\newblock \emph{Proceedings of the National Academy of Sciences}, 116\penalty0 (23):\penalty0 11537--11546, June 2019.
\newblock ISSN 0027-8424, 1091-6490.
\newblock \doi{10.1073/pnas.1820226116}.
\newblock URL \url{http://arxiv.org/abs/1810.10531}.
\newblock arXiv:1810.10531 [cs, q-bio, stat].

\bibitem[Boursier et~al.(2022)Boursier, Pillaud-Vivien, and Flammarion]{boursier_gradient_2022}
Etienne Boursier, Loucas Pillaud-Vivien, and Nicolas Flammarion.
\newblock Gradient flow dynamics of shallow {ReLU} networks for square loss and orthogonal inputs.
\newblock \emph{Advances in Neural Information Processing Systems}, 35:\penalty0 20105--20118, December 2022.
\newblock URL \url{https://proceedings.neurips.cc/paper_files/paper/2022/hash/7eeb9af3eb1f48e29c05e8dd3342b286-Abstract-Conference.html}.

\bibitem[Miyagawa(2023)]{miyagawa_toward_2023}
Taiki Miyagawa.
\newblock Toward {Equation} of {Motion} for {Deep} {Neural} {Networks}: {Continuous}-time {Gradient} {Descent} and {Discretization} {Error} {Analysis}, February 2023.
\newblock URL \url{http://arxiv.org/abs/2210.15898}.
\newblock arXiv:2210.15898 [cs].

\bibitem[Bertsekas(2000)]{bertsekas_dynamic_2000}
Dimitri~P. Bertsekas.
\newblock \emph{Dynamic {Programming} and {Optimal} {Control}}.
\newblock Athena Scientific, 2nd edition, October 2000.
\newblock ISBN 978-1-886529-09-0.

\bibitem[Boscain et~al.(2021)Boscain, Sigalotti, and Sugny]{boscain_introduction_2021}
U.~Boscain, M.~Sigalotti, and D.~Sugny.
\newblock Introduction to the {Pontryagin} {Maximum} {Principle} for {Quantum} {Optimal} {Control}.
\newblock \emph{PRX Quantum}, 2\penalty0 (3):\penalty0 030203, September 2021.
\newblock \doi{10.1103/PRXQuantum.2.030203}.
\newblock URL \url{https://link.aps.org/doi/10.1103/PRXQuantum.2.030203}.
\newblock Publisher: American Physical Society.

\bibitem[Chacko et~al.(2024)Chacko, P.c., and Abraham]{chacko_optimizing_2024}
Sanjay~Joseph Chacko, Neeraj P.c., and Rajesh~Joseph Abraham.
\newblock Optimizing {LQR} controllers: {A} comparative study.
\newblock \emph{Results in Control and Optimization}, 14:\penalty0 100387, March 2024.
\newblock ISSN 2666-7207.
\newblock \doi{10.1016/j.rico.2024.100387}.
\newblock URL \url{https://www.sciencedirect.com/science/article/pii/S2666720724000171}.

\bibitem[Piray and Daw(2021{\natexlab{a}})]{piray_linear_2021}
Payam Piray and Nathaniel~D. Daw.
\newblock Linear reinforcement learning in planning, grid fields, and cognitive control.
\newblock \emph{Nature Communications}, 12\penalty0 (1):\penalty0 4942, August 2021{\natexlab{a}}.
\newblock ISSN 2041-1723.
\newblock \doi{10.1038/s41467-021-25123-3}.
\newblock URL \url{https://www.nature.com/articles/s41467-021-25123-3}.
\newblock Number: 1 Publisher: Nature Publishing Group.

\bibitem[Bogacz et~al.(2006)Bogacz, Brown, Moehlis, Holmes, and Cohen]{bogacz_physics_2006}
Rafal Bogacz, Eric Brown, Jeff Moehlis, Philip Holmes, and Jonathan~D. Cohen.
\newblock The physics of optimal decision making: {A} formal analysis of models of performance in two-alternative forced-choice tasks.
\newblock \emph{Psychological Review}, 113\penalty0 (4):\penalty0 700--765, 2006.
\newblock ISSN 1939-1471.
\newblock \doi{10.1037/0033-295X.113.4.700}.
\newblock Place: US Publisher: American Psychological Association.

\bibitem[Bellman(1958)]{bellman_dynamic_1958}
Richard Bellman.
\newblock Dynamic programming and stochastic control processes.
\newblock \emph{Information and Control}, 1\penalty0 (3):\penalty0 228--239, September 1958.
\newblock ISSN 0019-9958.
\newblock \doi{10.1016/S0019-9958(58)80003-0}.
\newblock URL \url{https://www.sciencedirect.com/science/article/pii/S0019995858800030}.

\bibitem[Goldt et~al.(2019)Goldt, Advani, Saxe, Krzakala, and Zdeborová]{goldt_dynamics_2019}
Sebastian Goldt, Madhu Advani, Andrew~M Saxe, Florent Krzakala, and Lenka Zdeborová.
\newblock Dynamics of stochastic gradient descent for two-layer neural networks in the teacher-student setup.
\newblock In \emph{Advances in {Neural} {Information} {Processing} {Systems}}, volume~32. Curran Associates, Inc., 2019.
\newblock URL \url{https://proceedings.neurips.cc/paper_files/paper/2019/hash/cab070d53bd0d200746fb852a922064a-Abstract.html}.

\bibitem[Lee et~al.(2022)Lee, Mannelli, Clopath, Goldt, and Saxe]{lee_maslows_2022}
Sebastian Lee, Stefano~Sarao Mannelli, Claudia Clopath, Sebastian Goldt, and Andrew Saxe.
\newblock Maslow's {Hammer} for {Catastrophic} {Forgetting}: {Node} {Re}-{Use} vs {Node} {Activation}, May 2022.
\newblock URL \url{http://arxiv.org/abs/2205.09029}.
\newblock arXiv:2205.09029.

\bibitem[Deng(2012)]{deng_mnist_2012}
Li~Deng.
\newblock The {MNIST} {Database} of {Handwritten} {Digit} {Images} for {Machine} {Learning} {Research} [{Best} of the {Web}].
\newblock \emph{IEEE Signal Processing Magazine}, 29\penalty0 (6):\penalty0 141--142, November 2012.
\newblock ISSN 1558-0792.
\newblock \doi{10.1109/MSP.2012.2211477}.
\newblock URL \url{https://ieeexplore.ieee.org/document/6296535/citations}.

\bibitem[Horner(1993)]{horner_dynamics_1993}
Heinz Horner.
\newblock Dynamics of learning and generalization in perceptrons with constraints.
\newblock \emph{Physica A: Statistical Mechanics and its Applications}, 200\penalty0 (1):\penalty0 552--562, November 1993.
\newblock ISSN 0378-4371.
\newblock \doi{10.1016/0378-4371(93)90560-Q}.
\newblock URL \url{https://www.sciencedirect.com/science/article/pii/037843719390560Q}.

\bibitem[Saad and Solla(1995)]{saad_-line_1995}
David Saad and Sara~A. Solla.
\newblock On-line learning in soft committee machines.
\newblock \emph{Physical Review E}, 52\penalty0 (4):\penalty0 4225--4243, October 1995.
\newblock \doi{10.1103/PhysRevE.52.4225}.
\newblock URL \url{https://link.aps.org/doi/10.1103/PhysRevE.52.4225}.
\newblock Publisher: American Physical Society.

\bibitem[Ganjefar and Rezaei(2016)]{ganjefar_modified_2016}
Soheil Ganjefar and Sara Rezaei.
\newblock Modified homotopy perturbation method for optimal control problems using the {Padé} approximant.
\newblock \emph{Applied Mathematical Modelling}, 40\penalty0 (15):\penalty0 7062--7081, August 2016.
\newblock ISSN 0307-904X.
\newblock \doi{10.1016/j.apm.2016.02.039}.
\newblock URL \url{https://www.sciencedirect.com/science/article/pii/S0307904X16301147}.

\bibitem[Barari et~al.(2008)Barari, Omidvar, Ghotbi, and Ganji]{barari_application_2008}
A.~Barari, M.~Omidvar, Abdoul~R. Ghotbi, and D.~D. Ganji.
\newblock Application of {Homotopy} {Perturbation} {Method} and {Variational} {Iteration} {Method} to {Nonlinear} {Oscillator} {Differential} {Equations}.
\newblock \emph{Acta Applicandae Mathematicae}, 104\penalty0 (2):\penalty0 161--171, November 2008.
\newblock ISSN 1572-9036.
\newblock \doi{10.1007/s10440-008-9248-9}.
\newblock URL \url{https://doi.org/10.1007/s10440-008-9248-9}.

\bibitem[Abdulaziz et~al.(2008)Abdulaziz, Hashim, and Momani]{abdulaziz_application_2008}
O.~Abdulaziz, I.~Hashim, and S.~Momani.
\newblock Application of homotopy-perturbation method to fractional {IVPs}.
\newblock \emph{Journal of Computational and Applied Mathematics}, 216\penalty0 (2):\penalty0 574--584, July 2008.
\newblock ISSN 0377-0427.
\newblock \doi{10.1016/j.cam.2007.06.010}.
\newblock URL \url{https://www.sciencedirect.com/science/article/pii/S0377042707003081}.

\bibitem[Buhe et~al.(2023)Buhe, Rafiullah, Jabeen, and Anjum]{buhe_application_2023}
Eerdun Buhe, Muhammad Rafiullah, Dure Jabeen, and Naveed Anjum.
\newblock Application of homotopy perturbation method to solve a nonlinear mathematical model of depletion of forest resources.
\newblock \emph{Frontiers in Physics}, 11, October 2023.
\newblock ISSN 2296-424X.
\newblock \doi{10.3389/fphy.2023.1246884}.
\newblock URL \url{https://www.frontiersin.org/journals/physics/articles/10.3389/fphy.2023.1246884/full}.
\newblock Publisher: Frontiers.

\bibitem[Braun et~al.(2022)Braun, Dominé, Fitzgerald, and Saxe]{braun_exact_2022}
Lukas Braun, Clémentine Carla~Juliette Dominé, James~E. Fitzgerald, and Andrew~M. Saxe.
\newblock Exact learning dynamics of deep linear networks with prior knowledge.
\newblock In \emph{Advances in {Neural} {Information} {Processing} {Systems}}, October 2022.
\newblock URL \url{https://openreview.net/forum?id=lJx2vng-KiC}.

\bibitem[Bandura(1977)]{bandura_self-efficacy_1977}
Albert Bandura.
\newblock Self-efficacy: {Toward} a unifying theory of behavioral change.
\newblock \emph{Psychological Review}, 84\penalty0 (2):\penalty0 191--215, 1977.
\newblock ISSN 1939-1471.
\newblock \doi{10.1037/0033-295X.84.2.191}.
\newblock Place: US Publisher: American Psychological Association.

\bibitem[Zimmerman(2000)]{zimmerman_attaining_2000}
Barry~J. Zimmerman.
\newblock Attaining {Self}-{Regulation}.
\newblock In \emph{Handbook of {Self}-{Regulation}}, pages 13--39. Elsevier, 2000.
\newblock ISBN 978-0-12-109890-2.
\newblock \doi{10.1016/B978-012109890-2/50031-7}.
\newblock URL \url{https://linkinghub.elsevier.com/retrieve/pii/B9780121098902500317}.

\bibitem[Doya(2002)]{doya_metalearning_2002}
Kenji Doya.
\newblock Metalearning and neuromodulation.
\newblock \emph{Neural Networks}, 15\penalty0 (4):\penalty0 495--506, June 2002.
\newblock ISSN 0893-6080.
\newblock \doi{10.1016/S0893-6080(02)00044-8}.
\newblock URL \url{https://www.sciencedirect.com/science/article/pii/S0893608002000448}.

\bibitem[Aston-Jones and Cohen(2005)]{aston-jones_integrative_2005}
Gary Aston-Jones and Jonathan~D. Cohen.
\newblock An integrative theory of locus coeruleus-norepinephrine function: adaptive gain and optimal performance.
\newblock \emph{Annual Review of Neuroscience}, 28:\penalty0 403--450, 2005.
\newblock ISSN 0147-006X.
\newblock \doi{10.1146/annurev.neuro.28.061604.135709}.

\bibitem[Choi et~al.(2020)Choi, Shallue, Nado, Lee, Maddison, and Dahl]{choi_empirical_2020}
Dami Choi, Christopher~J. Shallue, Zachary Nado, Jaehoon Lee, Chris~J. Maddison, and George~E. Dahl.
\newblock On {Empirical} {Comparisons} of {Optimizers} for {Deep} {Learning}, June 2020.
\newblock URL \url{http://arxiv.org/abs/1910.05446}.
\newblock arXiv:1910.05446 [cs].

\bibitem[Schmidt et~al.(2021)Schmidt, Schneider, and Hennig]{schmidt_descending_2021}
Robin~M. Schmidt, Frank Schneider, and Philipp Hennig.
\newblock Descending through a {Crowded} {Valley} - {Benchmarking} {Deep} {Learning} {Optimizers}, August 2021.
\newblock URL \url{http://arxiv.org/abs/2007.01547}.
\newblock arXiv:2007.01547 [cs].

\bibitem[Kashyap(2023)]{kashyap_survey_2023}
Rohan Kashyap.
\newblock A survey of deep learning optimizers -- first and second order methods, September 2023.
\newblock URL \url{http://arxiv.org/abs/2211.15596}.
\newblock arXiv:2211.15596 [cs].

\bibitem[Pajares and Miller(1994)]{pajares_role_1994}
Frank Pajares and M.~David Miller.
\newblock Role of self-efficacy and self-concept beliefs in mathematical problem solving: {A} path analysis.
\newblock \emph{Journal of Educational Psychology}, 86\penalty0 (2):\penalty0 193--203, 1994.
\newblock ISSN 1939-2176.
\newblock \doi{10.1037/0022-0663.86.2.193}.
\newblock Place: US Publisher: American Psychological Association.

\bibitem[Zimmerman and Schunk(2011)]{zimmerman_self-regulated_2011}
Barry~J. Zimmerman and Dale~H. Schunk.
\newblock Self-{Regulated} {Learning} and {Performance}: {An} {Introduction} and an {Overview}.
\newblock In \emph{Handbook of {Self}-{Regulation} of {Learning} and {Performance}}. Routledge, 2011.
\newblock ISBN 978-0-203-83901-0.
\newblock Num Pages: 12.

\bibitem[Kontas and Özcan(2022)]{kontas_explaining_2022}
Hakki Kontas and Bahadir Özcan.
\newblock Explaining {Middle} {School} {Students}' {Mathematical} {Literacy} with {Sources} of {Self}-{Efficacy}, {Achievement} {Expectation} from {Family}, {Peers} and {Teachers}.
\newblock \emph{International Journal of Education and Literacy Studies}, 10\penalty0 (1):\penalty0 198--206, 2022.
\newblock URL \url{https://eric.ed.gov/?id=EJ1329200}.
\newblock Publisher: Australian International Academic Centre PTY, LTD ERIC Number: EJ1329200.

\bibitem[Son and Sethi(2006)]{son_metacognitive_2006}
Lisa~K. Son and Rajiv Sethi.
\newblock Metacognitive {Control} and {Optimal} {Learning}.
\newblock \emph{Cognitive Science}, 30\penalty0 (4):\penalty0 759--774, 2006.
\newblock ISSN 1551-6709.
\newblock \doi{10.1207/s15516709cog0000_74}.
\newblock URL \url{https://onlinelibrary.wiley.com/doi/abs/10.1207/s15516709cog0000_74}.
\newblock \_eprint: https://onlinelibrary.wiley.com/doi/pdf/10.1207/s15516709cog0000\_74.

\bibitem[Kool et~al.(2010)Kool, McGuire, Rosen, and Botvinick]{kool_decision_2010}
Wouter Kool, Joseph~T. McGuire, Zev~B. Rosen, and Matthew~M. Botvinick.
\newblock Decision making and the avoidance of cognitive demand.
\newblock \emph{Journal of Experimental Psychology: General}, 139\penalty0 (4):\penalty0 665--682, 2010.
\newblock ISSN 1939-2222.
\newblock \doi{10.1037/a0020198}.
\newblock Place: US Publisher: American Psychological Association.

\bibitem[Sweller(1988)]{sweller_cognitive_1988}
John Sweller.
\newblock Cognitive load during problem solving: {Effects} on learning.
\newblock \emph{Cognitive Science}, 12\penalty0 (2):\penalty0 257--285, 1988.
\newblock ISSN 1551-6709.
\newblock \doi{10.1207/s15516709cog1202_4}.
\newblock Place: Netherlands Publisher: Elsevier Science.

\bibitem[van Merriënboer and Sweller(2005)]{van_merrienboer_cognitive_2005}
Jeroen J.~G. van Merriënboer and John Sweller.
\newblock Cognitive {Load} {Theory} and {Complex} {Learning}: {Recent} {Developments} and {Future} {Directions}.
\newblock \emph{Educational Psychology Review}, 17\penalty0 (2):\penalty0 147--177, 2005.
\newblock ISSN 1573-336X.
\newblock \doi{10.1007/s10648-005-3951-0}.
\newblock Place: Germany Publisher: Springer.

\bibitem[Bandura and Schunk(1981)]{bandura_cultivating_1981}
Albert Bandura and Dale~H. Schunk.
\newblock Cultivating competence, self-efficacy, and intrinsic interest through proximal self-motivation.
\newblock \emph{Journal of Personality and Social Psychology}, 41\penalty0 (3):\penalty0 586--598, 1981.
\newblock ISSN 1939-1315.
\newblock \doi{10.1037/0022-3514.41.3.586}.
\newblock Place: US Publisher: American Psychological Association.

\bibitem[Schunk(1990)]{schunk_goal_1990}
Dale~H. Schunk.
\newblock Goal setting and self-efficacy during self-regulated learning.
\newblock \emph{Educational Psychologist}, 25\penalty0 (1):\penalty0 71--86, 1990.
\newblock ISSN 1532-6985.
\newblock \doi{10.1207/s15326985ep2501_6}.
\newblock Place: US Publisher: Lawrence Erlbaum.

\bibitem[Stock and Cervone(1990)]{stock_proximal_1990}
Jennifer Stock and Daniel Cervone.
\newblock Proximal goal-setting and self-regulatory processes.
\newblock \emph{Cognitive Therapy and Research}, 14\penalty0 (5):\penalty0 483--498, October 1990.
\newblock ISSN 1573-2819.
\newblock \doi{10.1007/BF01172969}.
\newblock URL \url{https://doi.org/10.1007/BF01172969}.

\bibitem[Rice(2001)]{rice_explaining_2001}
Jennifer~King Rice.
\newblock Explaining the {Negative} {Impact} of the {Transition} from {Middle} to {High} {School} on {Student} {Performance} in {Mathematics} and {Science}.
\newblock \emph{Educational Administration Quarterly}, 37\penalty0 (3):\penalty0 372--400, August 2001.
\newblock ISSN 0013-161X.
\newblock \doi{10.1177/00131610121969352}.
\newblock URL \url{https://doi.org/10.1177/00131610121969352}.
\newblock Publisher: SAGE Publications Inc.

\bibitem[Neild(2009)]{neild_falling_2009}
Ruth~Curran Neild.
\newblock Falling {Off} {Track} during the {Transition} to {High} {School}: {What} {We} {Know} and {What} {Can} {Be} {Done}.
\newblock \emph{The Future of Children}, 19\penalty0 (1):\penalty0 53--76, 2009.
\newblock ISSN 1054-8289.
\newblock URL \url{https://www.jstor.org/stable/27795035}.
\newblock Publisher: Princeton University.

\bibitem[Evans et~al.(2018)Evans, Borriello, and Field]{evans_review_2018}
Danielle Evans, Giulia~A. Borriello, and Andy~P. Field.
\newblock A {Review} of the {Academic} and {Psychological} {Impact} of the {Transition} to {Secondary} {Education}.
\newblock \emph{Frontiers in Psychology}, 9, August 2018.
\newblock ISSN 1664-1078.
\newblock \doi{10.3389/fpsyg.2018.01482}.
\newblock URL \url{https://www.frontiersin.org/journals/psychology/articles/10.3389/fpsyg.2018.01482/full}.
\newblock Publisher: Frontiers.

\bibitem[Hills(1965)]{hills_transfer_1965}
John~R. Hills.
\newblock Transfer {Shock}: {The} {Academic} {Performance} of the {Junior} {College} {Transfer}.
\newblock \emph{The Journal of Experimental Education}, 33\penalty0 (3):\penalty0 201--215, March 1965.
\newblock ISSN 0022-0973, 1940-0683.
\newblock \doi{10.1080/00220973.1965.11010875}.
\newblock URL \url{http://www.tandfonline.com/doi/abs/10.1080/00220973.1965.11010875}.

\bibitem[Lin et~al.(2023)Lin, Mastrokoukou, Longobardi, Bozzato, Gastaldi, and Berchiatti]{lin_students_2023}
Shanyan Lin, Sofia Mastrokoukou, Claudio Longobardi, Paolo Bozzato, Francesca Giovanna~Maria Gastaldi, and Martina Berchiatti.
\newblock Students' transition into higher education: {The} role of self-efficacy, regulation strategies, and academic achievements.
\newblock \emph{Higher Education Quarterly}, 77\penalty0 (1):\penalty0 121--137, 2023.
\newblock ISSN 1468-2273.
\newblock \doi{10.1111/hequ.12374}.
\newblock URL \url{https://onlinelibrary.wiley.com/doi/abs/10.1111/hequ.12374}.
\newblock \_eprint: https://onlinelibrary.wiley.com/doi/pdf/10.1111/hequ.12374.

\bibitem[Salmela-Aro et~al.(2008)Salmela-Aro, Kiuru, and Nurmi]{salmela-aro_role_2008}
Katariina Salmela-Aro, Noona Kiuru, and Jari-Erik Nurmi.
\newblock The role of educational track in adolescents' school burnout: {A} longitudinal study.
\newblock \emph{British Journal of Educational Psychology}, 78\penalty0 (4):\penalty0 663--689, 2008.
\newblock ISSN 2044-8279.
\newblock \doi{10.1348/000709908X281628}.
\newblock URL \url{https://onlinelibrary.wiley.com/doi/abs/10.1348/000709908X281628}.
\newblock \_eprint: https://bpspsychub.onlinelibrary.wiley.com/doi/pdf/10.1348/000709908X281628.

\bibitem[Barbosa et~al.(2016)Barbosa, Silva, Ferreira, and Severo]{barbosa_transition_2016}
Joselina Barbosa, Álvaro Silva, Maria~Amélia Ferreira, and Milton Severo.
\newblock Transition from {Secondary} {School} to {Medical} {School}: {The} {Role} of {Self}-{Study} and {Self}-{Regulated} {Learning} {Skills} in {Freshman} {Burnout}.
\newblock \emph{Acta Médica Portuguesa}, 29\penalty0 (12):\penalty0 803--808, December 2016.
\newblock ISSN 1646-0758.
\newblock \doi{10.20344/amp.8350}.
\newblock URL \url{https://www.actamedicaportuguesa.com/revista/index.php/amp/article/view/8350}.

\bibitem[Schunk and Rice(1991)]{schunk_learning_1991}
Dale~H. Schunk and Jo~M. Rice.
\newblock Learning goals and progress feedback during reading comprehension instruction.
\newblock \emph{Journal of Reading Behavior}, 23\penalty0 (3):\penalty0 351--364, 1991.
\newblock ISSN 0022-4111.
\newblock Place: US Publisher: Lawrence Erlbaum.

\bibitem[Glimcher(2011)]{glimcher_understanding_2011}
Paul~W. Glimcher.
\newblock Understanding dopamine and reinforcement learning: {The} dopamine reward prediction error hypothesis.
\newblock \emph{Proceedings of the National Academy of Sciences}, 108\penalty0 (supplement\_3):\penalty0 15647--15654, September 2011.
\newblock \doi{10.1073/pnas.1014269108}.
\newblock URL \url{https://www.pnas.org/doi/10.1073/pnas.1014269108}.
\newblock Publisher: Proceedings of the National Academy of Sciences.

\bibitem[Schultz(2016)]{schultz_dopamine_2016}
Wolfram Schultz.
\newblock Dopamine reward prediction error coding.
\newblock \emph{Dialogues in Clinical Neuroscience}, 18\penalty0 (1):\penalty0 23--32, March 2016.
\newblock ISSN 1294-8322.
\newblock \doi{10.31887/DCNS.2016.18.1/wschultz}.
\newblock URL \url{https://pmc.ncbi.nlm.nih.gov/articles/PMC4826767/}.

\bibitem[Taira and Sharpe(2025)]{taira_complementary_2025}
Masakazu Taira and Melissa~J Sharpe.
\newblock Complementary roles of serotonin and dopamine in model-based learning.
\newblock \emph{Current Opinion in Behavioral Sciences}, 61:\penalty0 101464, February 2025.
\newblock ISSN 2352-1546.
\newblock \doi{10.1016/j.cobeha.2024.101464}.
\newblock URL \url{https://www.sciencedirect.com/science/article/pii/S2352154624001153}.

\bibitem[Miyazaki et~al.(2014)Miyazaki, Miyazaki, Tanaka, Yamanaka, Takahashi, Tabuchi, and Doya]{miyazaki_optogenetic_2014}
Kayoko~W. Miyazaki, Katsuhiko Miyazaki, Kenji~F. Tanaka, Akihiro Yamanaka, Aki Takahashi, Sawako Tabuchi, and Kenji Doya.
\newblock Optogenetic activation of dorsal raphe serotonin neurons enhances patience for future rewards.
\newblock \emph{Current biology: CB}, 24\penalty0 (17):\penalty0 2033--2040, September 2014.
\newblock ISSN 1879-0445.
\newblock \doi{10.1016/j.cub.2014.07.041}.

\bibitem[Miyazaki et~al.(2018)Miyazaki, Miyazaki, Yamanaka, Tokuda, Tanaka, and Doya]{miyazaki_reward_2018}
Katsuhiko Miyazaki, Kayoko~W. Miyazaki, Akihiro Yamanaka, Tomoki Tokuda, Kenji~F. Tanaka, and Kenji Doya.
\newblock Reward probability and timing uncertainty alter the effect of dorsal raphe serotonin neurons on patience.
\newblock \emph{Nature Communications}, 9\penalty0 (1):\penalty0 2048, June 2018.
\newblock ISSN 2041-1723.
\newblock \doi{10.1038/s41467-018-04496-y}.
\newblock URL \url{https://www.nature.com/articles/s41467-018-04496-y}.
\newblock Publisher: Nature Publishing Group.

\bibitem[Meyniel et~al.(2016)Meyniel, Goodwin, Deakin, Klinge, MacFadyen, Milligan, Mullings, Pessiglione, and Gaillard]{meyniel_specific_2016}
Florent Meyniel, Guy~M. Goodwin, Jf~William Deakin, Corinna Klinge, Christine MacFadyen, Holly Milligan, Emma Mullings, Mathias Pessiglione, and Raphaël Gaillard.
\newblock A specific role for serotonin in overcoming effort cost.
\newblock \emph{eLife}, 5:\penalty0 e17282, November 2016.
\newblock ISSN 2050-084X.
\newblock \doi{10.7554/eLife.17282}.

\bibitem[Varazzani et~al.(2015)Varazzani, San-Galli, Gilardeau, and Bouret]{varazzani_noradrenaline_2015}
Chiara Varazzani, Aurore San-Galli, Sophie Gilardeau, and Sebastien Bouret.
\newblock Noradrenaline and {Dopamine} {Neurons} in the {Reward}/{Effort} {Trade}-{Off}: {A} {Direct} {Electrophysiological} {Comparison} in {Behaving} {Monkeys}.
\newblock \emph{Journal of Neuroscience}, 35\penalty0 (20):\penalty0 7866--7877, May 2015.
\newblock ISSN 0270-6474, 1529-2401.
\newblock \doi{10.1523/JNEUROSCI.0454-15.2015}.
\newblock URL \url{https://www.jneurosci.org/content/35/20/7866}.
\newblock Publisher: Society for Neuroscience Section: Articles.

\bibitem[Poe et~al.(2020)Poe, Foote, Eschenko, Johansen, Bouret, Aston-Jones, Harley, Manahan-Vaughan, Weinshenker, Valentino, Berridge, Chandler, Waterhouse, and Sara]{poe_locus_2020}
Gina~R. Poe, Stephen Foote, Oxana Eschenko, Joshua~P. Johansen, Sebastien Bouret, Gary Aston-Jones, Carolyn~W. Harley, Denise Manahan-Vaughan, David Weinshenker, Rita Valentino, Craig Berridge, Daniel~J. Chandler, Barry Waterhouse, and Susan~J. Sara.
\newblock Locus coeruleus: a new look at the blue spot.
\newblock \emph{Nature Reviews Neuroscience}, 21\penalty0 (11):\penalty0 644--659, November 2020.
\newblock ISSN 1471-0048.
\newblock \doi{10.1038/s41583-020-0360-9}.
\newblock URL \url{https://www.nature.com/articles/s41583-020-0360-9}.
\newblock Publisher: Nature Publishing Group.

\bibitem[Gilzenrat et~al.(2010)Gilzenrat, Nieuwenhuis, Jepma, and Cohen]{gilzenrat_pupil_2010}
Mark~S. Gilzenrat, Sander Nieuwenhuis, Marieke Jepma, and Jonathan~D. Cohen.
\newblock Pupil diameter tracks changes in control state predicted by the adaptive gain theory of locus coeruleus function.
\newblock \emph{Cognitive, Affective \& Behavioral Neuroscience}, 10\penalty0 (2):\penalty0 252--269, May 2010.
\newblock ISSN 1531-135X.
\newblock \doi{10.3758/CABN.10.2.252}.

\bibitem[Lee et~al.(2024)Lee, Garcia, Clopath, and Dabney]{lee_lifelong_2024}
Sebastian Lee, Samuel~Liebana Garcia, Claudia Clopath, and Will Dabney.
\newblock Lifelong {Reinforcement} {Learning} via {Neuromodulation}, August 2024.
\newblock URL \url{http://arxiv.org/abs/2408.08446}.
\newblock arXiv:2408.08446 [cs].

\bibitem[Flesch et~al.(2022)Flesch, Nagy, Saxe, and Summerfield]{flesch_modelling_2022}
Timo Flesch, David~G. Nagy, Andrew Saxe, and Christopher Summerfield.
\newblock Modelling continual learning in humans with {Hebbian} context gating and exponentially decaying task signals, March 2022.
\newblock URL \url{http://arxiv.org/abs/2203.11560}.
\newblock arXiv:2203.11560 [cs, q-bio].

\bibitem[Lillicrap and Kording(2019)]{lillicrap_what_2019}
Timothy~P. Lillicrap and Konrad~P. Kording.
\newblock What does it mean to understand a neural network?
\newblock \emph{arXiv:1907.06374 [cs, q-bio, stat]}, July 2019.
\newblock URL \url{http://arxiv.org/abs/1907.06374}.
\newblock arXiv: 1907.06374.

\bibitem[Song et~al.(2024)Song, Millidge, Salvatori, Lukasiewicz, Xu, and Bogacz]{song_inferring_2024}
Yuhang Song, Beren Millidge, Tommaso Salvatori, Thomas Lukasiewicz, Zhenghua Xu, and Rafal Bogacz.
\newblock Inferring neural activity before plasticity as a foundation for learning beyond backpropagation.
\newblock \emph{Nature Neuroscience}, 27\penalty0 (2):\penalty0 348--358, February 2024.
\newblock ISSN 1546-1726.
\newblock \doi{10.1038/s41593-023-01514-1}.
\newblock URL \url{https://www.nature.com/articles/s41593-023-01514-1}.
\newblock Publisher: Nature Publishing Group.

\bibitem[Cohen et~al.(1990)Cohen, Dunbar, and McClelland]{cohen_control_1990}
Jonathan~D. Cohen, Kevin Dunbar, and James~L. McClelland.
\newblock On the control of automatic processes: {A} parallel distributed processing account of the {Stroop} effect.
\newblock \emph{Psychological Review}, 97\penalty0 (3):\penalty0 332--361, 1990.
\newblock ISSN 1939-1471.
\newblock \doi{10.1037/0033-295X.97.3.332}.
\newblock Place: US Publisher: American Psychological Association.

\bibitem[Musslick et~al.(2020)Musslick, Saxe, Hoskin, Sagiv, Reichman, Petri, and Cohen]{musslick_rational_2020}
Sebastian Musslick, Andrew Saxe, Abigail Hoskin, Yotam Sagiv, Daniel Reichman, Giovanni Petri, and Jonathan Cohen.
\newblock On the {Rational} {Boundedness} of {Cognitive} {Control}: {Shared} {Versus} {Separated} {Representations}, November 2020.
\newblock URL \url{https://osf.io/jkhdf_v1}.

\bibitem[Pinon et~al.(2022)Pinon, Delvenne, and Jungers]{pinon_model-based_2022}
Brieuc Pinon, Jean-Charles Delvenne, and Raphaël Jungers.
\newblock A model-based approach to meta-{Reinforcement} {Learning}: {Transformers} and tree search, August 2022.
\newblock URL \url{http://arxiv.org/abs/2208.11535}.
\newblock arXiv:2208.11535 [cs].

\bibitem[Dong et~al.(2025)Dong, Yang, Liang, Feng, and Ran]{dong_adalrs_2025}
Hongyuan Dong, Dingkang Yang, Xiao Liang, Chao Feng, and Jiao Ran.
\newblock {AdaLRS}: {Loss}-{Guided} {Adaptive} {Learning} {Rate} {Search} for {Efficient} {Foundation} {Model} {Pretraining}, June 2025.
\newblock URL \url{http://arxiv.org/abs/2506.13274}.
\newblock arXiv:2506.13274 [cs].

\bibitem[Augustine(2023)]{augustine_note_2023}
Midhun~T. Augustine.
\newblock A {Note} on {Linear} {Quadratic} {Regulator} and {Kalman} {Filter}, August 2023.
\newblock URL \url{http://arxiv.org/abs/2308.15798}.
\newblock arXiv:2308.15798 [math].

\bibitem[Piray and Daw(2021{\natexlab{b}})]{piray_model_2021}
Payam Piray and Nathaniel~D. Daw.
\newblock A model for learning based on the joint estimation of stochasticity and volatility.
\newblock \emph{Nature Communications}, 12\penalty0 (1):\penalty0 6587, November 2021{\natexlab{b}}.
\newblock ISSN 2041-1723.
\newblock \doi{10.1038/s41467-021-26731-9}.
\newblock URL \url{https://www.nature.com/articles/s41467-021-26731-9}.
\newblock Publisher: Nature Publishing Group.

\bibitem[Si et~al.(2025)Si, Niu, and Wang]{si_review_2025}
Jiaqian Si, Yanxiong Niu, and Botao Wang.
\newblock A {Review} of {Nonlinear} {Filtering} {Algorithms} in {Integrated} {Navigation} {Systems}.
\newblock \emph{Sensors (Basel, Switzerland)}, 25\penalty0 (20):\penalty0 6462, October 2025.
\newblock ISSN 1424-8220.
\newblock \doi{10.3390/s25206462}.
\newblock URL \url{https://pmc.ncbi.nlm.nih.gov/articles/PMC12567677/}.

\bibitem[Yong and Zhou(1999)]{yong_dynamic_1999}
Jiongmin Yong and Xun~Yu Zhou.
\newblock Dynamic {Programming} and {HJB} {Equations}.
\newblock In Jiongmin Yong and Xun~Yu Zhou, editors, \emph{Stochastic {Controls}: {Hamiltonian} {Systems} and {HJB} {Equations}}, pages 157--215. Springer, New York, NY, 1999.
\newblock ISBN 978-1-4612-1466-3.
\newblock \doi{10.1007/978-1-4612-1466-3_4}.
\newblock URL \url{https://doi.org/10.1007/978-1-4612-1466-3_4}.

\bibitem[Beard et~al.(1998)Beard, Saridis, and Wen]{beard_approximate_1998}
R.~W. Beard, G.~N. Saridis, and J.~T. Wen.
\newblock Approximate {Solutions} to the {Time}-{Invariant} {Hamilton}–{Jacobi}–{Bellman} {Equation}.
\newblock \emph{Journal of Optimization Theory and Applications}, 96\penalty0 (3):\penalty0 589--626, March 1998.
\newblock ISSN 1573-2878.
\newblock \doi{10.1023/A:1022664528457}.
\newblock URL \url{https://doi.org/10.1023/A:1022664528457}.

\bibitem[Saberi~Nik et~al.(2012)Saberi~Nik, Effati, and Shirazian]{saberi_nik_approximate-analytical_2012}
H.~Saberi~Nik, S.~Effati, and M.~Shirazian.
\newblock An approximate-analytical solution for the {Hamilton}–{Jacobi}–{Bellman} equation via homotopy perturbation method.
\newblock \emph{Applied Mathematical Modelling}, 36\penalty0 (11):\penalty0 5614--5623, November 2012.
\newblock ISSN 0307-904X.
\newblock \doi{10.1016/j.apm.2012.01.013}.
\newblock URL \url{https://www.sciencedirect.com/science/article/pii/S0307904X12000285}.

\end{thebibliography}

\end{document}